%% file: main.tex
\documentclass{article}

\usepackage[preprint]{neurips_2026}

\usepackage[utf8]{inputenc}
\usepackage[T1]{fontenc}
\usepackage{hyperref}
\usepackage{url}
\usepackage{booktabs}
\usepackage{array}
\usepackage{amsfonts}
\usepackage{amsmath}
\usepackage{amssymb}
\usepackage{nicefrac}
\usepackage{microtype}
\usepackage{xcolor}
\usepackage{enumitem}
\usepackage{graphicx}
\usepackage[most]{tcolorbox}
\usepackage{alltt}

\definecolor{appendixgray}{HTML}{4A4A4A}
\definecolor{appendixbg}{HTML}{F7F7F7}
\newcolumntype{P}[1]{>{\raggedright\arraybackslash}p{#1}}

\newtcolorbox{promptbox}[1][]{
  enhanced,
  colback=appendixbg,
  colframe=appendixgray,
  boxrule=0.8pt,
  arc=2mm,
  left=2mm,
  right=2mm,
  top=1.5mm,
  bottom=1.5mm,
  title=#1,
  coltitle=white,
  colbacktitle=appendixgray,
  fonttitle=\bfseries,
}

\newcommand{\benchmarkname}{Agent\textsuperscript{2} RL-Bench}
\newcommand{\scorepos}[2]{#1\,{\scriptsize\textcolor{green!50!black}{(+#2)}}}
\newcommand{\scoreneg}[2]{#1\,{\scriptsize\textcolor{red!70!black}{($-$#2)}}}
\newcommand{\scorezero}[1]{#1\,{\scriptsize\textcolor{gray}{(0.00)}}}

\title{\benchmarkname: Can LLM Agents Engineer Agentic RL Post-Training?}

\author{%
  \textbf{Wanyi Chen}$^{1,2,3}$$^*$ \enspace
  \textbf{Xiao Yang}$^{3}$$^*$ \enspace
  \textbf{Xu Yang}$^{3}$$^*$ \enspace
  \textbf{Tianming Sha}$^{3,4}$$^*$ \enspace
  \textbf{Qizheng Li}$^{3,5}$ \\
  \textbf{Zhuo Wang}$^{3,6}$ \enspace
  \textbf{Bowen Xian}$^{3}$ \enspace
  \textbf{Fang Kong}$^{1,2}$$^\dagger$ \enspace
  \textbf{Weiqing Liu}$^{3}$$^\dagger$ \enspace
  \textbf{Jiang Bian}$^{3}$ \\
  {\normalfont $^1$School of Computer Science and Technology, Soochow University} \\
  {\normalfont $^2$Jiangsu Key Lab of Language Computing, Suzhou 215123, China} \\
  {\normalfont $^3$Microsoft Research Asia \quad $^4$Stony Brook University} \\
  {\normalfont $^5$Peking University \quad $^6$The University of Chicago}
}

\begin{document}

\maketitle

\begingroup
\renewcommand{\thefootnote}{}
\footnotetext{$^*$ Equal contribution. $^\dagger$ Corresponding authors. Benchmark contact: \texttt{berryccc0904@gmail.com}}
\endgroup

\input{sections/00_abstract}
\input{sections/01_introduction}
\input{sections/02_benchmark}
\input{sections/03_experiments}
\input{sections/04_related_work}
\input{sections/05_conclusion}

\begin{ack}
{This work was supported by the Project 62276178 under the National Natural Science Foundation of China, the Key Project 23KJA520012 under the Natural Science Foundation of Jiangsu Higher Education Institutions, and the Priority Academic Program Development of Jiangsu Higher Education Institutions. We thank Microsoft Research Asia for providing GPU resources and API access for this work. The authors declare no additional competing interests.}
\end{ack}

\bibliographystyle{plainnat}
\bibliography{references}

\clearpage
\appendix
\input{sections/06_appendix}

\end{document}

%% file: sections/00_abstract.tex
\begin{abstract}
We introduce \benchmarkname{}, a compact diagnostic benchmark for evaluating agentic RL post-training, which tests whether LLM agents can autonomously design, implement, debug, and execute post-training pipelines that improve foundation models. RL post-training increasingly drives model alignment and specialization, yet existing benchmarks are largely static, rewarding supervised fine-tuning or script generation without assessing an agent's ability to close an interactive RL loop. \benchmarkname{} provides a unified agent-facing interface: each run starts from an isolated workspace containing a base model, task data, instructions, and a grading API, and agents must iterate within a fixed budget by training models and submitting artifacts for evaluation. The benchmark spans six tasks across three levels, from static rule-based training to judge-based optimization and closed-loop online RL with trajectory collection. Two diagnostic skills, namely runtime recording and post-hoc summarization, enable structured analysis of agent behavior, facilitating smooth and effective iteration of the benchmark's evaluation framework. Across five agent systems and six driver LLMs, agents show intelligent behavior but clear limitations: one RL-oriented run improves ALFWorld from 4.85 to 93.28 via SFT warm-up and GRPO with online rollouts, yet DeepSearchQA remains difficult, most successful routes rely on supervised pipelines, and interactive outcomes show large  {differences} across agent stacks. Overall, \benchmarkname{} shows that current agents can sometimes engineer online RL, but stable agent-driven RL post-training remains rare under fixed budgets. It also demonstrates that our benchmark provides a strong and effective evaluation framework for future research in this direction. Code is available at  {\url{https://github.com/microsoft/RD-Agent}}.
\end{abstract}

%% file: sections/01_introduction.tex
\section{Introduction}

RL post-training now shapes how frontier models are aligned and specialized \citep{ouyang2022instructgpt,deepseek2025r1}, but building a working pipeline still requires expert judgment across rewards, algorithms, stability, and online data collection. If LLM agents could automate this process end to end, they would become powerful tools for model development. But can they?

We call this capability \textbf{agentic RL post-training}: an LLM agent autonomously designs, implements, debugs, and runs a post-training pipeline, potentially including closed-loop RL, to improve a given foundation model under a limited engineering budget. This is not a coding task. An agent must understand the training objective, select a strategy, build the data pipeline, implement environment interaction and rollout collection where needed, manage trajectory-level rewards, diagnose failures, and iterate until the model improves. It is a long-horizon systems engineering challenge.

Existing benchmarks only partially test this capability. Agent benchmarks have grown in complexity, from code generation \citep{chen2021humaneval} to software engineering \citep{swebench2023} to ML pipeline automation \citep{mlebench2024} and post-training \citep{posttrainbench2026}.
These benchmarks partially or directly evaluate automated post-training capabilities; however, they remain inherently static, optimizing only selected components within expert-designed pipelines. They do not require agents to perform rollout, handle trajectory-level rewards, or sustain online data collection. As a result, this static formulation constrains their scope and leaves a critical gap in evaluating agents' ability to autonomously design post-training processes.

We introduce \benchmarkname{} to close this gap. In this benchmark, the evaluated agent system must design a post-training agent that handles the full post-training pipeline, reflecting the key motivation behind the name ``\benchmarkname{}''. The benchmark contains six tasks across three progressively harder levels. Level~1 uses static rule-based tasks (GSM8K, HumanEval) with deterministic verification, testing data construction and supervised training under favorable conditions. Level~2 introduces static judge-based rewards (AlpacaEval), requiring optimization against a non-deterministic reward signal. Level~3 requires full environment interaction, trajectory collection, and closed-loop online RL under a fixed 12-hour budget (ALFWorld, WebShop, DeepSearchQA). Each level adds a structural requirement that the previous level does not impose, making the compact suite diagnostic.

Beyond final scores, \benchmarkname{} is designed for behavioral diagnosis. The benchmark provides each auto-designed agent an isolated workspace and a grading API for iterative submission, together with two key capabilities. First, \emph{runtime instrumentation} records every submission, code revision, and model artifact throughout the agent's run. Second, a \emph{post-hoc analysis module} automatically generates structured case studies and run reports from these artifacts. Together, these enable automated diagnostic analysis of agent-driven post-training behavior, revealing not just whether agents succeed but how they search, fail, and adapt.

In controlled experiments with five agent systems {(OpenHands, OpenCode, Claude Code, Codex CLI, Gemini CLI)} spanning three scaffold families and six driver LLMs, we find that agents achieve uneven interactive gains. On ALFWorld, an RL-oriented agent run improves from 4.85 to 93.28 via SFT warm-up and GRPO with online rollouts, showing that agent-driven online RL is possible in at least some settings.

Our contributions are:
\begin{itemize}[leftmargin=*, itemsep=1pt, topsep=2pt]
\item \textbf{A benchmark environment for agentic RL post-training with a plugin-based extensibility interface.} \benchmarkname{} provides a unified agent-facing environment spanning the six tasks and three levels described above, with two diagnostic skills (runtime recording and post-hoc summarization) that make agent behavior inspectable at the route and failure-mode level. Benchmarks and agents are each registered as standalone plugin directories (\texttt{benchmarks/\textless name\textgreater/} and \texttt{agents/\textless name\textgreater/}): adding a new task or a new agent system requires creating only one directory of small scope, without modifying the core evaluator, server, or any other plugin.
\item \textbf{Findings enabled by the benchmark's full-stack protocol.} Across five systems and six drivers, the route-tracked, mode-controlled, multi-level protocol surfaces three patterns. (i)  ALFWorld has strict online RL among its best pipelines (RL-oriented: 4.85 to 93.28 via SFT warm-up and GRPO rollouts);  {matched-budget comparisons also show online-RL gains on WebShop, while} score improvement alone is not sufficient evidence of online-RL engineering. (ii) At fixed scaffold and driver, the best-to-worst operating-mode spread on ALFWorld exceeds 80pp (13.43 to 95.52), a confound invisible to score-only leaderboards. (iii) Weak stacks actively regress the base model (up to  {$-$45pp} on HumanEval), confirming bidirectional discrimination.
\end{itemize}

%% file: sections/02_benchmark.tex
\section{\benchmarkname: An Agentic RL Benchmark}

\begin{figure}[t]
\centering
\includegraphics[width=1.0\linewidth]{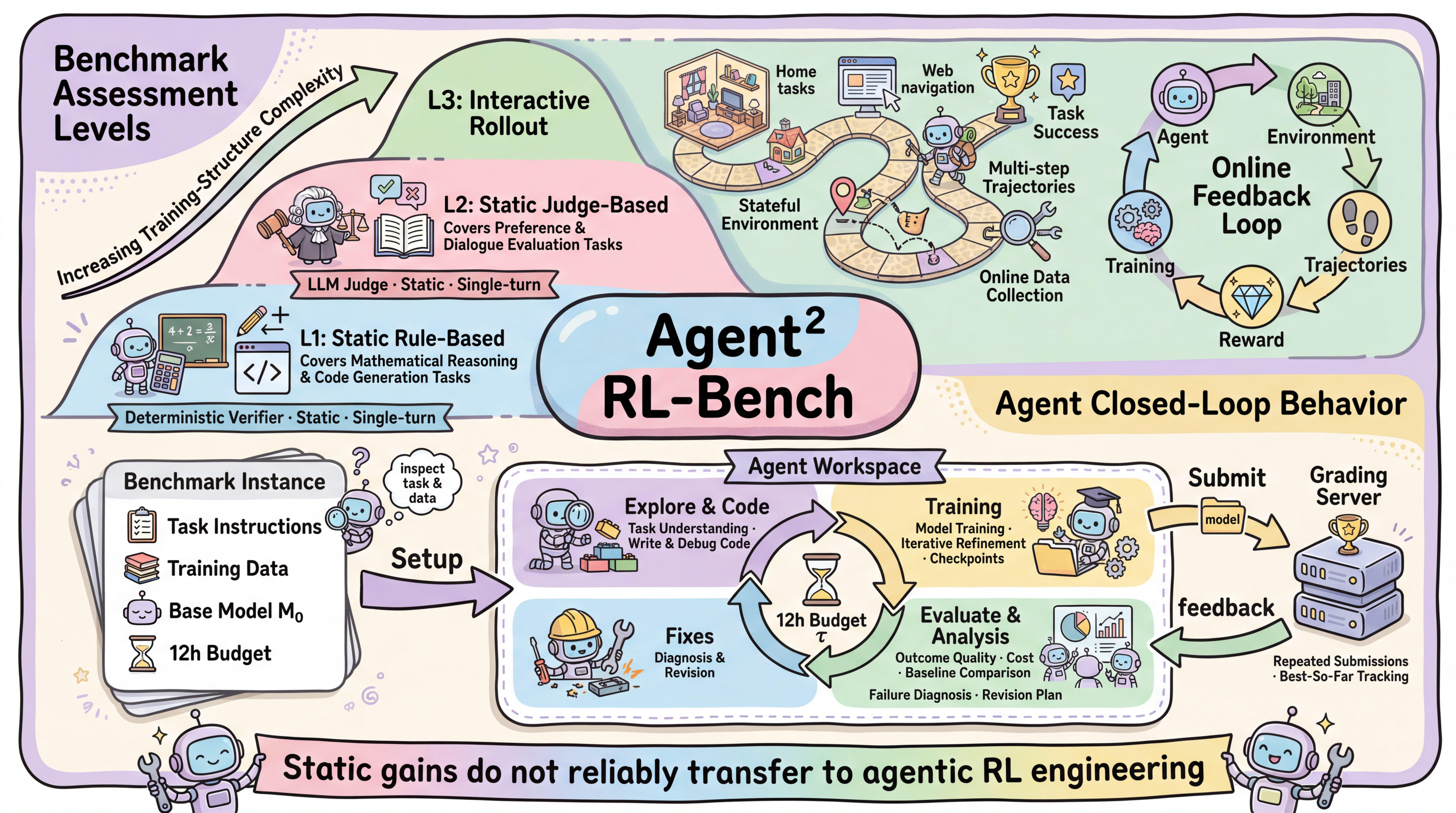}
\caption{Overview of \benchmarkname{}. Top-left: the three-level training-structure ladder, from static rule-based verification (L1), to static judge-based reward (L2), to interactive rollout with trajectory-level feedback (L3). Top-right: the L3 online feedback loop, where agents collect trajectories, receive rewards, and update models through environment interaction. Bottom: the shared run-level workflow: a benchmark instance (task instructions, training data, base model $M_0$, 12h budget) is placed in an isolated workspace; the agent writes code, trains models, submits artifacts, and receives scalar best-so-far feedback from the grading server.}
\label{fig:intro_overview}
\end{figure}

\subsection{Run-Level Evaluation Protocol}

Each benchmark instance is defined as $x=(\mathcal{D},M_0,\mathcal{T},\mathcal{O},\tau)$, consisting of a task description, base model, training data, evaluation oracle, and time budget. Within budget $\tau$, an LLM-driven agent $\mathcal{A}$ must inspect the task and data, choose a post-training route, write and execute training code, and submit candidate models to the grader. A run produces a time-ordered submission trace $\mathcal{R}(\mathcal{A},x)=\{(t_i,M_i,b_i,s_i)\}_{i=1}^{n}$, where $t_i$ is elapsed time, $M_i$ is the submitted model artifact, $b_i\in\{0,1\}$ indicates whether the submission is evaluable, and $s_i=\mathcal{O}(M_i)$ for valid submissions. The run is scored by the best valid submission:
\begin{equation}
\begin{aligned}
S(\mathcal{A},x)&=\max_{i:b_i=1}s_i,\\
\Delta(\mathcal{A},x)&=S(\mathcal{A},x)-\mathcal{O}(M_0),
\quad \operatorname{Succ}(\mathcal{A},x)={\mathbf{1}}[\Delta(\mathcal{A},x)>0].
\end{aligned}
\label{eq:run_protocol}
\end{equation}
If no valid submission is produced, the run receives the task-specific failure score.

This protocol evaluates the \emph{full post-training loop} rather than the quality of a generated training script in isolation. The best-of-budget score is deliberate: during real post-training work, diagnosis, repair, and resubmission are part of the engineering capability being evaluated.

\subsection{Training Structure Coverage}

\benchmarkname{} is organized around \emph{training structure coverage}. Rather than maximizing task count or topical diversity, we select tasks that span the structural transition from static post-training to full agentic RL engineering. Many existing post-training evaluations mostly measure whether an agent can assemble a useful fine-tuning workflow, whereas agentic RL becomes hardest when stateful interaction, trajectory handling, and online data generation are required. To make this distinction explicit, we organize the benchmark into three levels:

\begin{table}[t]
\centering
\caption{Benchmark overview organized by training structure.}
\label{tab:benchmark_overview}
\begin{tabular}{lllll}
\toprule
Level & Type & Benchmark & Reward & Horizon \\
\midrule
L1 & Static rule-based & GSM8K & Rule-based & Short \\
L1 & Static rule-based & HumanEval & Rule-based & Short \\
L2 & Static judge-based & AlpacaEval 2.0 & LLM judge & Short \\
L3 & Interactive rollout & ALFWorld & Sparse binary & Long \\
L3 & Interactive rollout & WebShop & Dense continuous & Long \\
L3 & Interactive rollout & DeepSearchQA & Judge-based & Medium \\
\bottomrule
\end{tabular}
\end{table}

L1 covers static rule-based post-training, where single-turn outputs are scored by deterministic verifiers. L2 introduces static judge-based post-training, where rewards are no longer directly programmable but the task remains single-turn and non-interactive. For both L1 and L2, evaluation reduces to independent single-turn outputs. L3 is the focus of the suite: interactive rollout tasks instead evaluate trajectories induced by the model. We write the two evaluator forms as
\begin{equation}
\begin{aligned}
\mathcal{O}_{\mathrm{static}}(M)
&=\frac{1}{N}\sum_{j=1}^{N} v_j(M(q_j)),\\
\xi_e(M)&=(o_1,a_1,r_1,\ldots,o_{H_e},a_{H_e},r_{H_e}),
\qquad
\mathcal{O}_{\mathrm{int}}(M)
=\frac{1}{N}\sum_{e=1}^{N}\phi_e(\xi_e(M)).
\end{aligned}
\label{eq:evaluator_forms}
\end{equation}
Here $v_j$ is either a deterministic verifier or a judge, and $\phi_e$ is a trajectory-level criterion: binary success for ALFWorld, task reward for WebShop, and search-and-judge correctness for DeepSearchQA.

The benchmark hypothesis is structural rather than topical: success on L1 or L2 should not be assumed to imply success on L3. In static settings, SFT often acts as a safe proxy for RL because reward is applied to a single output by a verifier or judge. In interactive settings, the agent must additionally implement environment stepping, maintain observation histories, handle trajectory-level rewards, and sustain online data collection. These requirements make L3 a meaningful test of agentic RL rather than post-training automation alone.

\subsection{Benchmark Selection}

We select six tasks to cover distinct capability types, reward structures, and decision horizons. GSM8K \citep{cobbe2021gsm8k} targets mathematical reasoning under a deterministic answer checker. HumanEval \citep{chen2021humaneval} targets code generation under rule-based unit tests. AlpacaEval 2.0 \citep{dubois2024alpacaeval} represents a static judge-based setting in which reward is not directly hand-coded. ALFWorld \citep{shridhar2020alfworld} and WebShop \citep{yao2022webshop} represent interactive long-horizon tasks with stateful environments. Our DeepSearchQA task is built on the \texttt{google/deepsearchqa} dataset \citep{deepsearchqa2026}, but uses a custom ReAct-style \citep{yao2023react} search-and-judge evaluator to fit the unified benchmark interface.

This benchmark set is intentionally compact. Rather than maximizing task count, we construct a small matrix in which each task contributes a distinct structural requirement: GSM8K and HumanEval anchor static rule-based post-training, AlpacaEval adds a non-programmable judge while remaining single-turn, ALFWorld and WebShop introduce stateful interaction under sparse and dense rewards, and DeepSearchQA adds tool use and external search. The intended use is therefore diagnostic rather than encyclopedic: the suite probes whether static post-training competence transfers to the online loop required by agentic RL systems. A four-dimensional structural-coverage visualization (capability breadth, interaction intensity, reward complexity, decision horizon) with the scoring rubric and task assignments is provided in Appendix~\ref{app:radar_rubric}.

\subsection{System Design}

Each run creates an isolated workspace containing read-only links to the base model and training data, together with writable directories for generated code and model outputs. The benchmark exposes a grading server that evaluates submissions, returns scores, and records the best result within the run. The repeated-submission design reflects that agentic RL typically depends on multiple cycles of training, diagnosis, and refinement rather than a single monolithic script.

The benchmark registry currently covers the six tasks described above. Static rule-based tasks are evaluated through OpenCompass-based pipelines. Judge-based tasks use task-specific evaluators. Interactive tasks rely on custom evaluators that support stateful environment interaction and rollout-based evaluation. The agent interface is fixed, while the evaluation backend varies with task structure. Figure~\ref{fig:intro_overview} maps these components onto the benchmark: the top panel summarizes the L1--L3 training-structure ladder, the right panel highlights the L3 online feedback loop, and the bottom panel shows the shared run-level workspace, submission, and grading workflow.

\subsection{Data Integrity and Evaluation}

Test sets are not mounted into the agent workspace, providing structural protection against leakage. The grading server returns only a scalar score and best-so-far, limiting but not eliminating adaptive overfitting from repeated submissions. In addition to $\Delta$ and $\operatorname{Succ}$, we derive process metrics from the trace: $\operatorname{ValidRate}(\mathcal{R})=\frac{1}{n}\sum_i b_i$, $t_{\mathrm{first}}=\min\{t_i:b_i=1,\ s_i>\mathcal{O}(M_0)\}$, and $t_{\mathrm{best}}=\min\{t_i:b_i=1,\ s_i=S(\mathcal{A},x)\}$. These metrics, together with submission count and offline route annotations, are recorded for diagnostic analysis. Details on evaluation isolation safeguards are in Appendix~\ref{app:data_integrity}. 

\subsection{Benchmark-Specific Details}

\paragraph{Static rule-based tasks.}
For GSM8K and HumanEval, the agent sees training data and task descriptions, while held-out evaluation is handled by the benchmark framework. GSM8K follows its standard train/test separation. HumanEval uses the standard 164 problems split disjointly by problem index into 82 training-visible and 82 held-out grading problems; absolute scores should be interpreted relative to this split rather than the original HumanEval leaderboard. Intended RL methods include single-turn methods such as GRPO or PPO over sampled completions, but the benchmark explicitly allows SFT as well. These tasks define the lower-structure end of the benchmark and test whether the agent can construct a useful post-training workflow under favorable conditions.

\paragraph{Static judge-based task.}
AlpacaEval 2.0 differs from rule-based settings because reward is mediated by a judge rather than a deterministic checker. It tests whether agents can optimize against a preference-like signal while remaining in a non-interactive regime, thereby bridging fully programmable rewards and interactive RL. In our implementation, this task preserves the single-turn structure of AlpacaEval while integrating it into the same repeated-submission loop as the other tasks.

\paragraph{Interactive rollout tasks.}
ALFWorld and WebShop require stateful multi-step interaction, while DeepSearchQA combines multi-step search with judge-based evaluation in a tool-augmented interactive loop. These tasks form the benchmark's primary stress test for agentic RL competence because useful training data must be grounded in task interaction or trajectory construction rather than plain single-turn fine-tuning alone. More broadly, the three tasks expose different failure modes: sparse-reward long-horizon interaction in ALFWorld, dense but brittle task-specific reward in WebShop, and tool use plus judge-mediated search in DeepSearchQA.

\paragraph{Task adaptations and evaluator reliability.}
\benchmarkname{} standardizes heterogeneous tasks under one agent-facing submission loop, so some wrappers are adaptations rather than exact reproductions: AlpacaEval preserves its single-turn preference structure but is exposed through the unified grading API, and DeepSearchQA uses the public \texttt{google/deepsearchqa} data with a custom ReAct-style search-and-judge evaluator. Evaluation protocols differ across tasks (deterministic verifiers for L1, an LLM judge for AlpacaEval, 134 fixed episodes for ALFWorld, 100 instructions for WebShop, and a held-out search-and-judge set for DeepSearchQA), so scores should be interpreted relative to each task's verifier and protocol rather than compared directly to original leaderboards.

%% file: sections/03_experiments.tex
\section{Experiments}

\subsection{Experimental Setup}
\label{sec:experimental_setup}

We organize experiments into two complementary studies. {For both studies, baseline scores are averaged over four evaluations.} The first is a \textbf{scaffold comparison} on Qwen2.5-7B-Instruct~\citep{qwen2024qwen25} with a uniform 12h budget: OpenHands and OpenCode, both driven by GPT-5.4, and Claude Code driven by Claude Opus 4.6 (four operating modes: SFT-only, RL-oriented, Free, Multi). Table~\ref{tab:main_results} reports  {Free-mode results in both panels} and Table~\ref{tab:mode_comparison} gives the full {single-run }mode ablation\textcolor{black}{.} {Appendix~\ref{app:campaign_cost} reports original single-run costs.}

The second is a \textbf{controlled CLI-agent study} on Qwen3-8B-Base~\citep{qwen2025qwen3} (unaligned), using a fixed 12h budget{ per run}: three CLI scaffolds (Claude Code, Codex~CLI, Gemini~CLI) paired with six driver LLMs (Claude Opus 4.6, Claude Sonnet~4.5, GPT-5.4, GPT-5.2, GPT-4o, gemini-2.5-flash), yielding  {six} system configurations. This partial factorial design enables within-scaffold driver comparisons that were not possible in the 7B study. The RL-oriented mode instructs the agent to prioritize RL or online reward optimization, while allowing SFT or behavior-cloning warm-up when needed for stability. Mode constraints in Claude Code are enforced via system-prompt instructions and verified through exhaustive post-hoc workspace audit (\texttt{grep} over all generated code files; results in Appendix~\ref{app:alfworld_audit}).

\textbf{Important caveat.} The two studies are complementary rather than fully factorial: the 7B study entangles scaffold and driver, while the 8B study only partially disentangles them through within-scaffold driver swaps.  {Observed} differences should be read as suggestive stack-level signals rather than isolated scaffold, driver, or seed effects. The most controlled comparison remains the mode ablation in Section~\ref{sec:mode_comparison}, where scaffold and driver are held fixed.

\subsection{Cross-Agent Comparison}

Table~\ref{tab:main_results} reports four-run mean scores; sample SDs are provided in Appendix~\ref{app:run_variability}.

\begin{table*}[t]
\centering
\caption{\benchmarkname{} leaderboard. All runs use a 12h budget.  {Agent results: mean over four runs.} \textit{Panel~A (7B-Instruct):} scaffold comparison.  \textit{Panel~B (8B-Base):} \enspace{scaffold and driver comparison} on Qwen3-8B-Base across three CLI scaffolds and six driver LLMs. {Avg $\Delta$ is the unweighted mean improvement over the corresponding baseline across six benchmarks (pp).}}
\label{tab:main_results}
\small
\setlength{\tabcolsep}{3.5pt}
\resizebox{\textwidth}{!}{
\begin{tabular}{llccccccc}
\toprule
& & \multicolumn{2}{c}{L1: Static Rule} & \multicolumn{1}{c}{L2: Judge} & \multicolumn{3}{c}{L3: Interactive Rollout} & \\
\cmidrule(lr){3-4}\cmidrule(lr){5-5}\cmidrule(lr){6-8}
System & Driver & GSM8K & HumanEval & AlpacaEval & ALFWorld & WebShop & DeepSearchQA & {\shortstack{Avg $\Delta$\\(pp) $\uparrow$}} \\
\midrule
\multicolumn{9}{l}{\textit{Panel A: Qwen2.5-7B-Instruct (scaffold comparison)}} \\
\midrule
Baseline & N/A & 77.70 & 75.61 & 18.90 & 4.85 & 69.00 & 11.75 & {0.00} \\
OpenHands & GPT-5.4 & 80.18 & 75.92 & 18.93 & 8.58 & 46.00 & 8.38 & {-3.31} \\
OpenCode & GPT-5.4 & 82.13 & 80.49 & 21.41 & 18.47 & 59.75 & \textbf{19.88} & {+4.05} \\
Claude Code & Opus 4.6 & \textbf{82.20} & \textbf{80.80} & \textbf{30.34} & \textbf{94.40} & \textbf{83.25} & 13.63 & {\textbf{+21.13}} \\
\midrule
\multicolumn{9}{l}{\textit{Panel B: Qwen3-8B-Base (\enspace{scaffold and driver comparison})}} \\
\midrule
Baseline & N/A & 83.02 & 62.20 & 14.68 & 8.96 & 0.00 & 12.25 & {0.00} \\
Claude Code & Opus 4.6 & 88.31 & \textbf{84.76} & \textbf{26.74} & \textbf{95.15} & \textbf{79.25} & \textbf{23.88} & {\textbf{+36.16}} \\
Claude Code & Sonnet 4.5 & \textbf{90.07} & 64.94 & 19.61 & 71.27 & 72.25 & 16.75 & {+25.63} \\
Codex CLI & GPT-5.4 & 84.38 & 71.65 & 22.54 & 88.43 & 53.75 & 22.50 & {+27.02} \\
Codex CLI & GPT-5.2 & 89.86 & 72.56 & 17.44 & 86.38 & 73.00 & 13.13 & {+28.54} \\
Codex CLI & GPT-4o & 84.59 & 63.41 & 16.46 & 10.26 & 25.50 & 15.75 & {+5.81} \\
Gemini CLI & 2.5-flash & 66.36 & 16.77 & 6.55 & 10.26 & 18.00 & 12.50 & {-8.44} \\
\bottomrule
\end{tabular}
}
\end{table*}

The two panels reveal complementary findings. In the 7B scaffold comparison (Panel~A), Claude Code achieves the  {highest mean scores} on 5 of 6 tasks under a uniform 12h budget, with particularly large gains on interactive tasks:  {+89.55} on ALFWorld and  {+14.25} on WebShop versus OpenCode's  {+13.62} and  {$-$9.25}, respectively. DeepSearchQA is the one task where OpenCode outperforms ( {+8.13} vs.\  {+1.88}), suggesting that the advantage is environment-dependent rather than a uniform scaffold ordering.

The controlled 8B study (Panel~B) shows \textbf{large observed agent-stack sensitivity on interactive tasks}. Within Codex~CLI, switching from GPT-4o to GPT-5.4 changes ALFWorld improvement from  {+1.30} to  {+79.47} using the same scaffold, with GPT-5.2 also reaching  {+77.42}. Within Claude Code, Opus reaches  {+86.19} on ALFWorld and  {+79.25} on WebShop, while Sonnet~4.5 reaches {+62.31} on ALFWorld with substantial run-to-run variability and {+72.25} on WebShop. Consistent with the setup caveat, we treat these as stack-level signals; the magnitude is task-dependent, with a  {24pp} span on GSM8K ( {66.36 to 90.07}) versus  {85pp} on ALFWorld ( {10.26 to 95.15}).

Three cross-cutting observations emerge. First, interactive success is not a single-system phenomenon: Claude Code + Opus, Codex~CLI + GPT-5.4, and Codex~CLI + GPT-5.2 all achieve strong ALFWorld gains, while Claude Code + Sonnet~4.5 and Codex~CLI + GPT-5.2 achieve strong WebShop gains. Second, DeepSearchQA remains structurally hard: the  {highest mean score} is  {23.88 (+11.63}, Claude Code + Opus), followed by  {22.50 (+10.25}, Codex~CLI + GPT-5.4), but absolute scores remain far below ALFWorld and WebShop. Third, the weakest stack (Gemini~CLI + 2.5-flash) \emph{regresses} below baseline by  {$-$16.66} pp on GSM8K and  {$-$45.43} pp on HumanEval.

\subsection{Behavioral Analysis: Capability Stratification}

\paragraph{Score improvement and online-RL engineering are separable.} A central evaluation lesson of \benchmarkname{} is that score improvement is not sufficient evidence of online-RL engineering. ALFWorld shows that agents can close the strict online loop, yet most {of the }highest-scoring routes are SFT-based or SFT-initialized, and even the best ALFWorld score uses a hybrid composite pipeline rather than strict GRPO. We therefore report route type as a diagnostic attribute: trajectory SFT shows that an agent can construct task-grounded data from environment interaction, while online RL shows that it can close the rollout-reward-update loop. {Matched-budget online RL outperforms trajectory SFT on ALFWorld and WebShop (Appendix~\ref{app:matched_online_rl}).}

Workspace artifacts show qualitatively distinct strategies across tasks, suggesting capability stratification rather than a static/interactive boundary: tasks sit at different stages of tractability that shift with scaffold quality (Appendix~\ref{app:cc_cases}; Figure~\ref{fig:mode_heatmap}).

\begin{figure*}[t]
\centering
\includegraphics[width=0.90\textwidth]{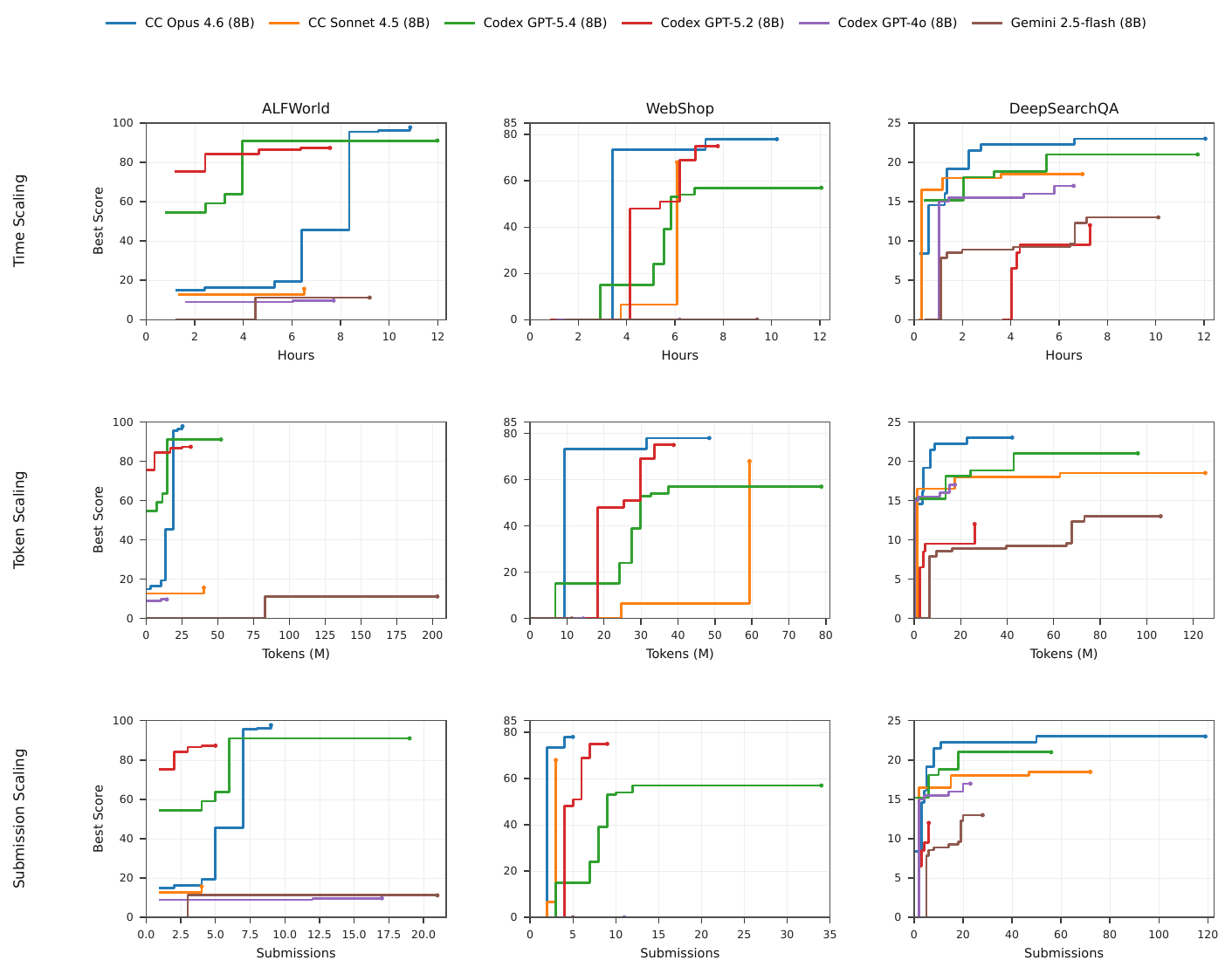}
\caption{Interactive-task scaling across agent stacks. Rows plot best-so-far score vs.\ elapsed hours, context tokens, and submissions; static-task scaling is in Appendix Figure~\ref{fig:static_scaling}.}
\label{fig:combined_scaling}
\vspace{-0.6em}
\end{figure*}

\paragraph{Cross-agent behavioral differences.}
OpenCode-7B illustrates the same boundary from a weaker-scaffold direction: on WebShop, it regressed by  {$-$9.25}, while Claude Code achieved  {+14.25} under the same 12h protocol. This sign reversal suggests that trajectory collection quality, not algorithm choice alone, is the primary bottleneck. Figure~\ref{fig:combined_scaling} shows that interactive tasks benefit most from late progress and separate agent stacks more sharply than static tasks. {Figures~\ref{fig:combined_scaling}, \ref{fig:static_scaling}, and~\ref{fig:scaffold_driver_heatmap} show original single-run results.}

\paragraph{ALFWorld near-saturation.} On ALFWorld, the RL-oriented mode improves from 4.85 to 93.28 via SFT warm-up followed by GRPO with online rollouts, the clearest case of online RL succeeding in our benchmark. The Free mode reaches 95.52 in the 7B {mode }study and 97.76 in the 8B Opus run through a composite supervised pipeline combining trajectory collection and training-evaluation format alignment, though this does not constitute online RL engineering in the strict sense (Appendix~\ref{app:alfworld_trace}).

\paragraph{DeepSearchQA remains structurally hard.} In the 7B mode study, the most exhaustive DeepSearchQA run produced 145 submissions and tried DPO~\citep{rafailov2023dpo}, GRPO, KTO, ORPO, and model merging, but reached only 15.0 (+3.25 over baseline).  {The 8B mean scores reach}  {23.88 (+11.63)} for Claude Code + Opus and  {22.50 (+10.25)} for Codex~CLI + GPT-5.4, yet absolute scores remain much lower than ALFWorld and WebShop. The task therefore exposes a different failure mode from ALFWorld: agents can construct elaborate post-training pipelines, but the search-retrieve-judge loop remains difficult to improve through post-training alone. For context, even frontier agents with full web access achieve only 66\% fully-correct on the official DeepSearchQA leaderboard~\citep{deepsearchqa2026}.

\begin{figure*}[t]
\centering
\includegraphics[width=\textwidth]{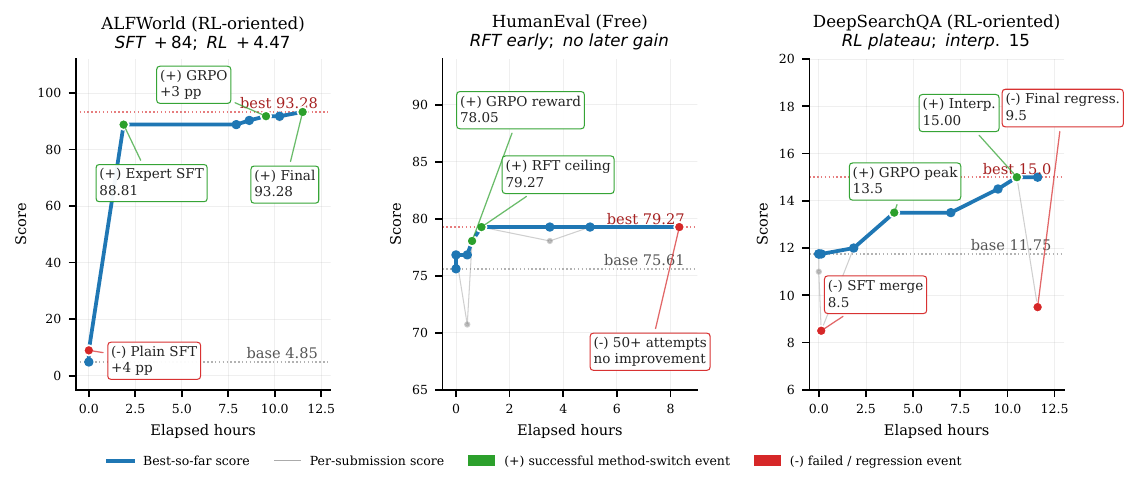}
\caption{Method-switch trajectories for three representative 7B-Instruct runs. Blue/gray show best-so-far/per-submission scores; green/red mark successful/failed route changes, showing discrete route pivots and engineering interventions rather than smooth hill-climbing.}
\label{fig:trajectory_methods}
\vspace{-0.6em}
\end{figure*}

\paragraph{Method attribution.} The workspace audit makes the route-type point concrete: across reconstructable Claude Code 7B cells, SFT is attempted in 19 and adopted in 18 best routes, while GRPO is attempted in 8 but adopted in only 2 (Table~\ref{tab:method_adoption}). Figure~\ref{fig:trajectory_methods} shows the same pattern at run level: score jumps align with route changes and engineering interventions rather than smooth hill-climbing. Static benchmark gains are weak predictors of interactive gains; Claude Code's ALFWorld improvement is more than an order of magnitude larger than its GSM8K improvement.

\subsection{Operating Mode Analysis}
\label{sec:mode_comparison}

Since Claude Code is the only scaffold with explicit operating-mode control, we use it to study the effect of constraining the training paradigm. Table~\ref{tab:mode_comparison} reports per-mode results on both 7B-Instruct (12h) and 8B-Base (12h, Multi vs.\ Free only).

\begin{table*}[t]
\centering
\caption{Single-run Claude Code mode ablation. Parenthetical values are absolute score changes from the corresponding base-model baseline. \textit{Top:} Qwen2.5-7B-Instruct (12h, Opus 4.6). \textit{Bottom:} Qwen3-8B-Base (12h, Opus 4.6, Free vs.\ Multi only). Bold marks the best score within each base-model block.}
\label{tab:mode_comparison}
\small
\setlength{\tabcolsep}{4.5pt}
\renewcommand{\scorepos}[2]{#1\,{\scriptsize (+#2)}}
\renewcommand{\scoreneg}[2]{#1\,{\scriptsize ($-$#2)}}
\renewcommand{\scorezero}[1]{#1\,{\scriptsize (0.00)}}
\resizebox{\textwidth}{!}{
\begin{tabular}{lcccccc}
\toprule
Mode & GSM8K & HumanEval & AlpacaEval & ALFWorld & WebShop & DeepSearchQA \\
\midrule
\multicolumn{7}{l}{\textit{7B-Instruct}} \\
\midrule
Baseline & 77.70 & 75.61 & 18.90 & 4.85 & 69.00 & 11.75 \\
SFT-only & \scorepos{84.00}{6.30} & \textbf{\scorepos{81.71}{6.10}} & \scorepos{21.32}{2.42} & \scorepos{67.91}{63.06} & \scorepos{76.00}{7.00} & \scorepos{14.00}{2.25} \\
RL-oriented & \scorepos{81.88}{4.18} & \textbf{\scorepos{81.71}{6.10}} & \scorepos{20.96}{2.06} & \scorepos{93.28}{88.43} & \scorepos{79.00}{10.00} & \textbf{\scorepos{15.00}{3.25}} \\
Free & \scorepos{82.26}{4.56} & \scorepos{79.27}{3.66} & \textbf{\scorepos{30.76}{11.86}} & \textbf{\scorepos{95.52}{90.67}} & \textbf{\scorepos{84.00}{15.00}} & \scorepos{13.00}{1.25} \\
Multi & \textbf{\scorepos{85.06}{7.36}} & \scorezero{75.61} & \scorepos{19.54}{0.64} & \scorepos{13.43}{8.58} & \scorepos{79.00}{10.00} & \scoreneg{11.50}{0.25} \\
\midrule
\multicolumn{7}{l}{\textit{8B-Base}} \\
\midrule
Baseline & 83.02 & 62.20 & 14.68 & 8.96 & 0.00 & 12.25 \\
Free & \shortstack{\\{\textbf{85.60\,{\scriptsize (+2.58)}}}} & \textbf{\scorepos{84.15}{21.95}} & \textbf{\scorepos{26.91}{12.23}} & \textbf{\scorepos{97.76}{88.80}} & \textbf{\scorepos{78.00}{78.00}} & \textbf{\scorepos{23.00}{10.75}} \\
Multi & {\scoreneg{80.97}{2.05}} & \scoreneg{60.98}{1.22} & \scoreneg{10.33}{4.35} & \scorepos{85.82}{76.86} & \scorepos{13.00}{13.00} & \scoreneg{7.00}{5.25} \\
\bottomrule
\end{tabular}
}
\end{table*}

The rank ordering of modes {in the 7B study } \emph{reverses} between L1 and L3 tasks: SFT-only is near-best on GSM8K but lags composite routes on WebShop and ALFWorld, while RL-oriented achieves 93.28 on ALFWorld but is weakest on GSM8K. Free mode produces the strongest ALFWorld, WebShop, and AlpacaEval results through composite strategies unavailable under single-paradigm constraints. Multi-task joint optimization wins only on GSM8K {in the 7B study }and severely degrades interactive tasks on both  {base models}. The best-to-worst mode spread exceeds 80 points on ALFWorld, showing that operating mode is a major hidden variable in benchmark reporting. Panel~B also suggests that unaligned starts expose alignment-gap closure as a stack-dependent outcome: strong stacks reach near- or above-Instruct performance on multiple tasks, whereas weaker stacks can erase baseline capability.

%% file: sections/04_related_work.tex
\section{Related Work}

LLM evaluation has moved from static capability suites such as BIG-bench~\citep{srivastava2022bigbench} and HELM~\citep{liang2022helm} toward more agentic settings: software-engineering benchmarks and agent interfaces (SWE-bench~\citep{swebench2023}, SWE-agent~\citep{yang2024sweagent}, OpenHands~\citep{wang2024openhands}), tool and web interaction (AgentBench~\citep{agentbench2023}, GAIA~\citep{gaia2023}, WebArena~\citep{webarena2023}), and ML experimentation or AI research automation (MLAgentBench~\citep{huang2024mlagentbench}, MLE-bench~\citep{mlebench2024}, MLGym~\citep{nathani2025mlgym}, RE-Bench~\citep{wijk2024rebench}).

Recent work on agent-driven post-training and autonomous fine-tuning includes PostTrainBench~\citep{posttrainbench2026}, which evaluates bounded post-training tasks, and FT-Dojo~\citep{li2026ftdojo}, which studies language agents for autonomous LLM fine-tuning. These settings largely remain static or expert-scaffolded, focusing on supervised or offline pipelines rather than rollout collection, trajectory-level rewards, and online data generation. \benchmarkname{} extends this line of evaluation to closed-loop RL and post-training under a fixed engineering budget.

Our work is complementary to RL post-training methods such as InstructGPT-style RLHF~\citep{ouyang2022instructgpt}, GRPO-style training in DeepSeekMath~\citep{shao2024deepseekmath}, and DeepSeek-R1~\citep{deepseek2025r1}: we evaluate autonomous engineering rather than proposing a new optimizer or training recipe.

%% file: sections/05_conclusion.tex
\section{Conclusion}

\benchmarkname{} is a compact diagnostic benchmark for agentic RL post-training, spanning static rule-based tasks, static judge-based rewards, and closed-loop online RL with trajectory collection. Our experiments show both promise and clear limits: agents can sometimes close interactive loops, yet DeepSearchQA remains difficult, most successful routes rely on supervised fine-tuning, and outcomes are stack-sensitive. Static post-training benchmarks therefore understate the offline-to-online engineering gap; score gains alone do not prove online-RL engineering, which remains rare under fixed budgets.

\paragraph{Limitations and Future Work.}
\benchmarkname{} is compute-intensive: the  {mode ablations use} single runs, {and the 8B study }omits many scaffold-driver pairs and covers only six tasks, leaving broader RL environments, reward designs, and model scales to future work. Full-stack evaluation also entangles planning, training stability, and driver quality. Future work should add repeated  {mode ablations}, broader online RL tasks, component-attribution ablations, calibration baselines, and hidden final evaluation more broadly with submission caps.

%% file: sections/06_appendix.tex
\section{Benchmark Protocol and Task Details}

The appendix follows the same order as the evidence burden of the paper. We first document the benchmark protocol, task contracts, metrics, and prompts. We then give route definitions and representative code evidence, followed by reproducibility and anti-cheating safeguards. The later sections provide behavioral diagnostics, additional quantitative results, and detailed trace walkthroughs. This organization separates benchmark-native scoring from auxiliary process analysis.

\subsection{Structural Coverage Rubric}
\label{app:radar_rubric}

\begin{figure}[!h]
\centering
\includegraphics[width=0.72\linewidth]{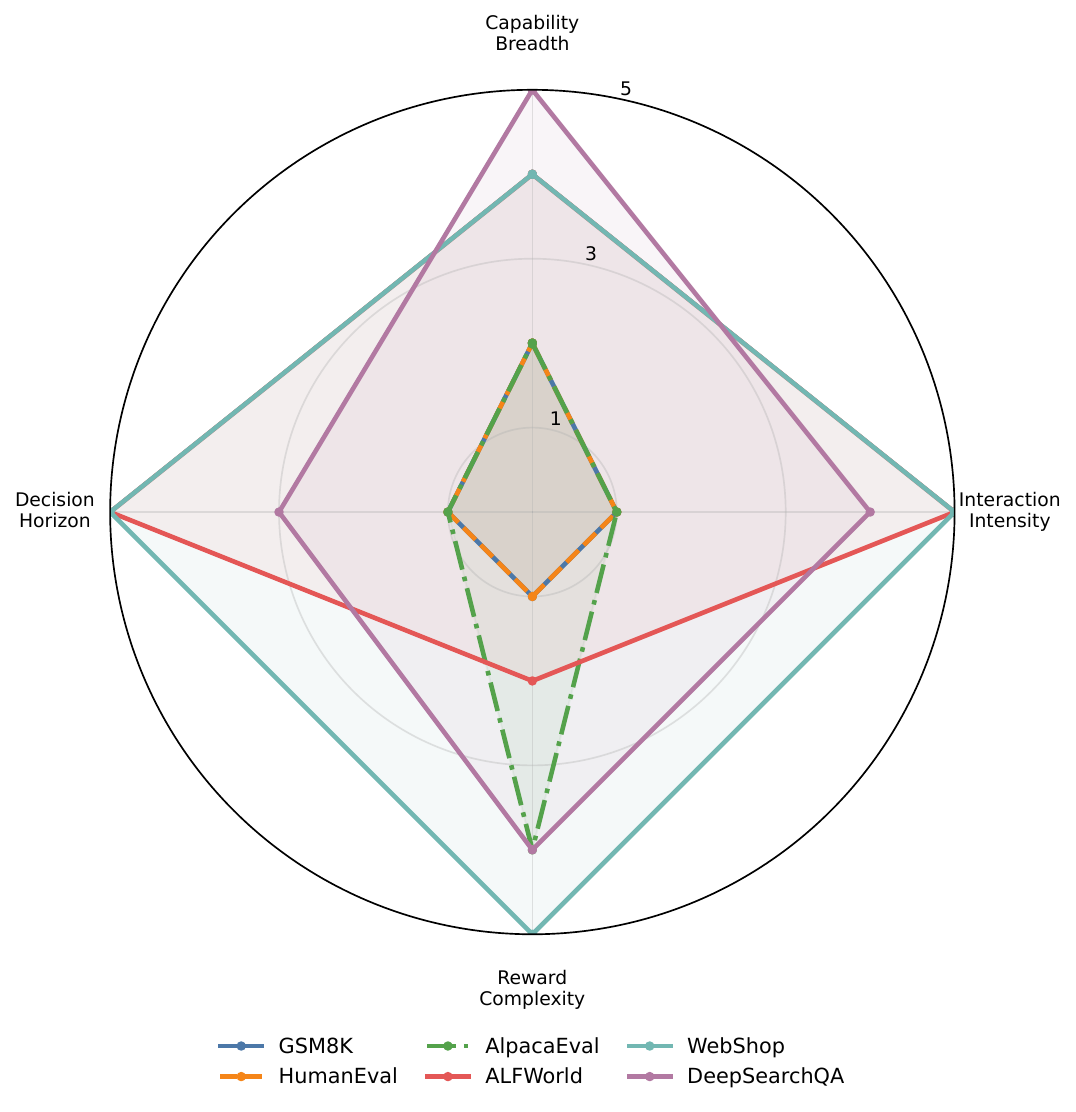}
\caption{Coverage of the six benchmark tasks across four manually defined structural dimensions: capability breadth, interaction intensity, reward complexity, and decision horizon. Scores are ordinal (1 to 5) and summarize benchmark-level metadata rather than empirical results.}
\label{fig:benchmark_coverage}
\end{figure}

Figure~\ref{fig:benchmark_coverage} is a benchmark-design visualization rather than an empirical result. Each benchmark is assigned an ordinal score from 1 to 5 on four manually defined structural dimensions. These scores are intended only to summarize structural differences between tasks and to make the benchmark coverage visually explicit.

\paragraph{Dimension definitions.}
\begin{itemize}
\item \textbf{Capability breadth}: how many qualitatively different abilities must be combined to solve the task. A score of 1 indicates a narrow, mostly single-capability task; a score of 5 indicates a task requiring broad coupling of reasoning, planning, search, tool use, or state tracking.
\item \textbf{Interaction intensity}: how strongly the task depends on explicit multi-step interaction with a stateful environment or iterative tool loop. A score of 1 indicates a static single-shot task; a score of 5 indicates strong dependence on repeated environment interaction.
\item \textbf{Reward complexity}: how difficult the reward signal is to specify or optimize against. Rule-based exact-match rewards are low; dense or judge-based rewards are higher.
\item \textbf{Decision horizon}: the effective length of the action or reasoning sequence needed to complete the task. Single-turn generation is low; long multi-step interaction is high.
\end{itemize}

\begin{table}[h]
\centering
\caption{Ordinal benchmark-design scores used in Figure~\ref{fig:benchmark_coverage}.}
\label{tab:radar_scores}
\begin{tabular}{lcccc}
\toprule
Benchmark & Capability & Interaction & Reward & Horizon \\
\midrule
GSM8K & 2 & 1 & 1 & 1 \\
HumanEval & 2 & 1 & 1 & 1 \\
AlpacaEval & 2 & 1 & 4 & 1 \\
ALFWorld & 4 & 5 & 2 & 5 \\
WebShop & 4 & 5 & 5 & 5 \\
DeepSearchQA & 5 & 4 & 4 & 3 \\
\bottomrule
\end{tabular}
\end{table}

\subsection{Benchmark Protocol Details}

Table~\ref{tab:appendix_task_protocol} summarizes the task-level evaluation contracts. The benchmark is unified at the outer loop, but the inner evaluator, reward source, and horizon differ substantially across tasks.

\begin{table*}[t]
\centering
\caption{Task protocol summary following the current implementation.}
\label{tab:appendix_task_protocol}
\small
\setlength{\tabcolsep}{4pt}
\resizebox{\textwidth}{!}{
\begin{tabular}{lcccccc}
\toprule
Task & Structure & Evaluator Type & Reward Source & Eval Size & Horizon & Primary Difficulty \\
\midrule
GSM8K & Static rule-based & OpenCompass & Exact-match answer check & standard test split & Short & reasoning-to-training transfer \\
HumanEval & Static rule-based & OpenCompass & unit-test execution & 82 held-out / 82 train-visible & Short & code-generation post-training \\
AlpacaEval 2.0 & Static judge-based & LLM judge & pairwise preference judging & reference split & Short & non-programmable reward \\
ALFWorld & Interactive rollout & environment evaluator & sparse success reward & 134 episodes & Long & rollout and trajectory collection \\
WebShop & Interactive rollout & environment evaluator & task-specific dense reward & 100 instructions & Long & reward shaping and stability \\
DeepSearchQA & Interactive tool-use & custom ReAct evaluator & search-and-judge loop & 200 samples & Medium & tool use, search, and judged QA \\
\bottomrule
\end{tabular}
}
\end{table*}

\paragraph{Static rule-based tasks.}
For GSM8K and HumanEval, the agent receives training data and task descriptions, while held-out evaluation is handled by the benchmark framework. GSM8K follows its standard train/test separation. HumanEval uses the standard 164 problems split into 82 training-visible problems and 82 held-out grading problems. These tasks permit SFT and single-turn RL methods and anchor the low-structure end of the suite.

\paragraph{Static judge-based tasks.}
AlpacaEval 2.0 occupies a middle point between static verification and interactive RL. The task remains single-turn, but the reward is supplied by an LLM judge rather than a deterministic checker.

\paragraph{Interactive rollout tasks.}
ALFWorld and WebShop require explicit interaction loops with stateful environments. DeepSearchQA is implemented as a tool-augmented interactive QA task built on the \texttt{google/deepsearchqa} dataset, using a custom ReAct-style search-and-judge evaluator rather than the original paper's full protocol. Across all three tasks, useful improvement requires task-grounded interaction or trajectory construction rather than plain single-turn fine-tuning alone. Scores for adapted tasks should be interpreted as benchmark-internal measurements under the unified agent interface, not as direct submissions to the original leaderboards.

\subsection{Reported Metrics}

We distinguish \emph{benchmark-native outputs} from \emph{offline diagnostics}. This distinction is important for interpreting the main results.

\begin{table}[h]
\centering
\caption{Reported metric types in \benchmarkname{}.}
\label{tab:appendix_metrics}
\begin{tabular}{lp{0.62\linewidth}}
\toprule
Metric Type & Definition \\
\midrule
Benchmark-native outputs & baseline score, per-submission score, best score, improvement over baseline, submission count, runtime metadata \\
Process metrics & valid-submission rate, time to first improvement, time to best, score trajectory, time-used ratio, last-submit gap \\
Offline diagnostics & route labels, coding-success indicators, case-study inspection, trace-based interpretation \\
\bottomrule
\end{tabular}
\end{table}

The benchmark-native outputs are produced directly by the grading-server loop and are the basis of the main quantitative claims. Process metrics are derived deterministically from saved run artifacts such as \texttt{scores.json} and run metadata. Offline diagnostics are used only for interpretation and should not be confused with the benchmark's native scoring interface.

\subsection{Experimental Setup Table}

\begin{table}[t]
\centering
\caption{Experimental setup used for the main experiments.}
\label{tab:setup}
\begin{tabular}{ll}
\toprule
Dimension & Configuration \\
\midrule
\multicolumn{2}{l}{\textit{7B Main Scaffold Comparison}} \\
\midrule
Agents & OpenHands, OpenCode, Claude Code \\
Driver models & GPT-5.4 (OH/OC), Claude Opus 4.6 (CC) \\
Base model & Qwen2.5-7B-Instruct \\
Time budget & 12h  {per run} \\
CC modes & SFT-only, RL-oriented, Free, Multi \\
\midrule
\multicolumn{2}{l}{\textit{8B Controlled CLI Study}} \\
\midrule
Scaffolds & Claude Code, Codex CLI, Gemini CLI \\
Driver models & Opus 4.6, Sonnet 4.5, GPT-5.4, GPT-5.2, GPT-4o, 2.5-flash \\
Base model & Qwen3-8B-Base (unaligned) \\
Time budget & 12h  {per run} \\
\midrule
\multicolumn{2}{l}{\textit{Shared Evaluation}} \\
\midrule
ALFWorld eval split & 134 episodes \\
WebShop eval size & 100 instructions \\
DeepSearchQA eval size & 200 held-out samples \\
Judge model & GPT-4o \\
\bottomrule
\end{tabular}
\end{table}

 {Each run has a 12h budget; Table~\ref{tab:main_results} reports four-run means (sample SDs in Appendix~\ref{app:run_variability}), and Table~\ref{tab:mode_comparison} reports single-run mode ablations.} We additionally inspected earlier exploratory Claude Code workspaces to understand route discovery and failure modes. These exploratory traces are labeled as such when discussed and are used only for qualitative diagnosis, not for the main leaderboard or scaffold comparison.

\subsection{Run Contract}

Each run in \benchmarkname{} is defined by the tuple \emph{(agent scaffold, task, base model, time budget)}. The outer-loop contract is deliberately simple. The runner prepares an isolated workspace, mounts the task-specific data and model artifacts, starts a grading server, computes the baseline score for the provided base model, and then grants the agent autonomous control within the workspace.

At execution time, the agent is given access to a small set of standard environment variables, including the task identifier, the base-model path, the workspace path, the output directory, and the grading-server endpoint \texttt{GRADING\_SERVER\_URL}. Submissions are made by posting a model path back to the grading server, which evaluates the submission, appends the result to \texttt{scores.json}, and returns the current score together with the best score observed so far.

\paragraph{Practical constraints.}
The run contract also includes a bounded engineering budget. Agents are not evaluated in an interactive chat setting with human feedback; instead, they are expected to operate autonomously until timeout or natural termination. In the current implementation, some scaffolds are additionally given a lightweight \texttt{timer.sh} utility so they can query the remaining time budget during long runs.

\paragraph{Submission granularity.}
Submission count should not be read as the number of full end-to-end training jobs. A single training run can yield multiple candidate checkpoints, merged variants, or adapter-merged artifacts, and many submissions are quick evaluations of existing artifacts rather than new full training runs. Some submissions also correspond to short LoRA/SFT runs on filtered data, failed attempts, or diagnostic variants. The submission trace therefore measures how often an agent queries the grader during the engineering loop, not how many times it trains a model from scratch.

\subsection{Agent Instruction Excerpt}

The benchmark also provides an agent-facing instruction file (\texttt{core/instructions.md}) that standardizes what every scaffold is told about the workspace, timing signals, and submission API. We do not reproduce the full file verbatim here; instead, we include the operative excerpt that defines the contract most relevant to reproducibility and auditability.

\begin{promptbox}[Agent Workspace and Submission Instructions]
\small
\textbf{Core goal.} Improve the score as much as possible within a fixed time budget, submit multiple times, and iterate using evaluation feedback.\\[0.5mm]
\textbf{Workspace restrictions.} The current directory is the workspace. Use only relative paths. Do not \texttt{cd} outside the workspace.\\[0.5mm]
\textbf{Time source of truth.} Read \texttt{run\_meta.json} first. The file records \texttt{start\_time}, \texttt{timeout\_s}, \texttt{last\_submit\_time}, and \texttt{end\_time}.\\[0.5mm]
\textbf{Environment variables.} \texttt{TASK}, \texttt{BASE\_MODEL}, \texttt{MODEL\_PATH}, \texttt{DATA\_PATH}, \texttt{OUTPUT\_DIR}, and \texttt{GRADING\_SERVER\_URL}.\\[0.5mm]
\textbf{Expected loop.}
\begin{enumerate}
\item Read \texttt{description.md}, \texttt{instructions.md}, and task-specific files such as \texttt{eval.py} when provided.
\item Write training code under \texttt{code/} and train a candidate model.
\item Save the trained model under \texttt{output/}.
\item Submit the candidate by posting \texttt{\{"model\_path": "..."\}} to \texttt{\$GRADING\_SERVER\_URL/submit}.
\item Use the returned score and best-so-far signal to decide the next iteration.
\end{enumerate}
\textbf{Submission note.} Submitting the untouched base model is explicitly discouraged and maps back to baseline performance; LoRA-based methods must merge adapters before submission.
\end{promptbox}

\subsection{Structured Run Reports}

In addition to benchmark-native outputs such as \texttt{scores.json} and run metadata, each run maintains two structured reporting artifacts under \texttt{reports/}. The first is \texttt{summary.md}, a human-readable cumulative report intended for manual inspection. The second is \texttt{summary.jsonl}, a compact machine-readable log with one record per iteration.

\paragraph{Human-readable summaries.}
The \texttt{summary.md} file is append-only and is designed to make long agent runs auditable without replaying the full terminal trace. Each iteration records the run status, score and improvement, inferred training type, key configuration choices, a short explanation of what was attempted, the rationale for that attempt, the main failure mode or progress signal, and a small code excerpt from the generated training script when available. In practice, this file is the main artifact we consult when reconstructing why a run failed, why a route changed, or why a particular submission improved over baseline.

\paragraph{Machine-readable iteration logs.}
The companion \texttt{summary.jsonl} file stores a minimal quantitative record for each iteration, including fields such as iteration index, elapsed-time marker, duration, status, exit code, score, improvement, training type, and failure type. We use this file for lightweight downstream aggregation, while leaving richer causal interpretation to \texttt{summary.md} and the saved workspace itself.

\paragraph{Role in the paper.}
These run reports are not part of the benchmark-native scoring interface and are never used to define the main leaderboard-style claims. Instead, they serve as auxiliary instrumentation for the offline analysis reported in Section~3, especially for route analysis, failure taxonomy, and case-study reconstruction. This separation is deliberate: the benchmark score should come only from the evaluator, whereas the run reports help us interpret how the agent reached that score.

\begin{promptbox}[Example Structured Run Report]
\small
\textbf{Human-readable report (\texttt{reports/summary.md})}\\[0.5mm]
\texttt{\#\# Iteration 4 (T+04:18, duration 812s)}\\
\texttt{- Status: success (exit\_code=0)}\\
\texttt{- Score: 46.21 | Improvement: +38.00 | Best: 46.21}\\
\texttt{- Train type: SFT+RL}\\
\texttt{- Key config: beta-mixed rollout collection}\\
\texttt{- Failure/progress signal: first positive submission}\\[1.5mm]
\textbf{Machine-readable log (\texttt{reports/summary.jsonl})}\\[0.5mm]
\texttt{\{"iteration": 4, "duration\_s": 812, "status": "success",}\\
\texttt{"score": 46.21, "improvement": 38.00, "train\_type": "SFT+RL",}\\
\texttt{"failure\_type": "unknown"\}}
\end{promptbox}

\subsection{Judge Prompt Excerpts}

Judge-based tasks are important in \benchmarkname{} because they make the reward signal less programmable than exact-match or unit-test metrics. We therefore include representative prompt excerpts here to clarify the form of supervision used in those tasks.

\paragraph{AlpacaEval-style preference judging.}
When the built-in AlpacaEval pipeline is unavailable or falls back to the project-level API backend, the judge compares two candidate answers to the same instruction and returns one of \texttt{A}, \texttt{B}, or \texttt{TIE}. The prompt is instantiated from the following template:

\begin{promptbox}[AlpacaEval-Style Judge Prompt]
\small
You are evaluating two candidate answers to the same instruction.\\
Instruction: \{instruction\}\\
Candidate A: \{candidate\}\\
Candidate B: \{reference\}\\
Reply with exactly one label: A, B, or TIE.
\end{promptbox}

\paragraph{DeepSearchQA answer judging.}
DeepSearchQA uses a task-specific answer judge rather than a pairwise preference annotator. The judge receives the answer type, the gold answer, and the predicted answer, and is instructed to return only \texttt{correct} or \texttt{incorrect}. The current implementation uses the following prompt structure:

\begin{promptbox}[DeepSearchQA Judge Prompt]
\small
You are an answer evaluator. Compare the predicted answer to the gold answer.\\
Question answer type: \{answer\_type\}\\
Gold answer: \{gold\}\\
Predicted answer: \{predicted\}\\
For Single Answer: the predicted answer is correct if it contains the same key information as the gold answer.\\
For Set Answer: the predicted answer is correct if it contains all items from the gold answer.\\
Reply with only ``correct'' or ``incorrect''.
\end{promptbox}

These excerpts are not the only ingredients in the full evaluation pipeline, but they illustrate the main difference between the two judge-based settings: AlpacaEval relies on preference comparison between two candidate responses, whereas DeepSearchQA relies on answer verification conditioned on a gold target and answer type.

\section{Route and Code Evidence}

The technical-route analysis in the main text is based on the generated \texttt{train.py} script in each saved workspace. Table~\ref{tab:appendix_route_defs} summarizes the labels used in that analysis.

\begin{table}[h]
\centering
\caption{Route labels used in the technical-route analysis.}
\label{tab:appendix_route_defs}
\begin{tabular}{lp{0.60\linewidth}}
\toprule
Route & Operational definition \\
\midrule
GRPO & \texttt{train.py} contains explicit GRPO-style training components \\
PPO & \texttt{train.py} contains explicit PPO-style training components \\
SFT-only & \texttt{train.py} contains supervised fine-tuning components but no explicit RL trainer \\
SFT+RL & \texttt{train.py} contains both supervised and RL-style training components \\
Copy & the run copies or reuses the base model without substantive training logic \\
Placeholder / Missing & the training script is absent or effectively a placeholder rather than executable training code \\
Coding success & a run that produces substantive training code, operationalized as GRPO, PPO, SFT-only, or SFT+RL \\
Improvement success & a run with at least one submission whose improvement over baseline is strictly positive \\
\bottomrule
\end{tabular}
\end{table}

This analysis is intentionally operational rather than semantic. Its purpose is not to recover the full intent of every generated script, but to separate runs that genuinely attempt training from runs that fail earlier and only produce superficial artifacts.

\subsection{Supplementary Case Studies}

\paragraph{OpenCode on ALFWorld.}
A representative positive ALFWorld case follows a three-stage structure: offline expert SFT, online beta-mixed DAgger~\citep{ross2011dagger} collection, and aggregated SFT on the resulting data. The run first submits a model below baseline and only later recovers, which makes it a useful example of late route discovery rather than steady incremental improvement. We therefore interpret it primarily as route-discovery evidence. To make the route auditable without relying on the exact historical workspace artifact, we also retain a cleaned reproduction variant of the same route shape and summarize its key components in Appendix~\ref{app:code_evidence}.

\paragraph{WebShop collapse.}
WebShop illustrates a different failure mode: a run may generate substantial code and still fail to produce useful learning signal. In the representative negative case, the training script implements oracle bootstrap trajectories, relaxed DAgger collection, explicit step-level reward shaping, and advantage-weighted behavioral cloning, yet the run still fails to improve over baseline and later produces a non-evaluable submission. This case supports the distinction between ``code exists'' and ``useful RL training happened.'' In interactive tasks, reward handling, trajectory quality, and training stability are part of the problem itself rather than secondary implementation details.

\subsection{Representative Code Evidence for RL and Hybrid Routes}
\label{app:code_evidence}

The following excerpts are included only as code-level evidence for route shape. They are not benchmark-native metrics, and they should not be interpreted as replacing the quantitative results in the main paper.

\paragraph{GSM8K: explicit GRPO route.}
One representative GSM8K run uses an explicit GRPO trainer with exact-match reward over extracted final answers:

\begin{promptbox}[GSM8K GRPO Fragment]
\footnotesize
\begin{alltt}
from trl import GRPOConfig, GRPOTrainer

def compute_reward(prompts, completions, **kwargs):
    answers = kwargs.get("answers", [])
    rewards = []
    for i, completion in enumerate(completions):
        answer = answers[i] if i < len(answers) else ""
        extracted = extract_answer(completion)
        answer_str = str(answer).strip()
        rewards.append(1.0 if extracted == answer_str else 0.0)
    return rewards

trainer = GRPOTrainer(
    model=model,
    args=training_args,
    train_dataset=train_dataset,
    eval_dataset=eval_dataset,
    reward_funcs=compute_reward,
    processing_class=tokenizer,
)
\end{alltt}
\end{promptbox}

This excerpt is useful because it shows that at least some static-task runs are not merely SFT pipelines: the agent can instantiate a genuine RL-style route when the reward is cleanly programmable.

\paragraph{ALFWorld: cleaned DAgger-style route.}
To make the discovered ALFWorld route auditable without depending on a task-specific prompt file, we distilled the route into a cleaned reproduction variant. The core mechanics remain expert imitation, beta-mixed online collection, shaped transition scoring, and replay-based retraining:

\begin{promptbox}[ALFWorld Prompt and Teacher Fragment]
\footnotesize
\begin{alltt}
def load_react_prompts(workspace: Path) -> Dict[str, str]:
    prompts_path = workspace / "react_prompts.json"
    if prompts_path.exists():
        return cast(Dict[str, str], load_json(prompts_path))
    fallback = {\{}f"react_{\{}key{\}}_fallback": default_react_prompt(key)
                for key in sorted(set(TASK_PREFIXES.values())){\}}
    print(f"[train] warning: {\{}prompts_path.name{\}} not found; "
          "using generic task-conditioned prefixes instead.")
    return fallback

def teacher_beta(iteration: int, cfg: TrainConfig) -> float:
    mix = iteration / float(cfg.dagger_iterations - 1)
    cosine_mix = 0.5 - 0.5 * math.cos(math.pi * mix)
    delta = cfg.teacher_beta_end - cfg.teacher_beta_start
    return cfg.teacher_beta_start + delta * cosine_mix
\end{alltt}
\end{promptbox}

\begin{promptbox}[ALFWorld Action-Scoring Fragment]
\footnotesize
\begin{alltt}
def score_candidate(action, recent_actions, observation,
                    admissible_commands, expert_action) -> float:
    score = 0.0
    observation_l = observation.lower()
    admissible_set = set(admissible_commands)
    recent_signatures = {\{}action_signature(x) for x in recent_actions{\}}
    sig = action_signature(action)

    if not is_valid_action(action):
        return -1.0e9
    if action in admissible_set:
        score += 2.0
    if action == expert_action:
        score += 1.5
    if sig and sig not in recent_signatures:
        score += 0.25
    if sig in recent_signatures:
        score -= 0.40
    if action.startswith("think:"):
        score -= 0.50
    if action in {\{}"look", "inventory"{\}}:
        score -= 0.20
    return score
\end{alltt}
\end{promptbox}

\begin{promptbox}[ALFWorld Expert-Demonstration Fragment for Qwen2.5-7B-Instruct Free Mode]
\footnotesize
\textbf{Exploratory note:} This code is from an earlier 24h Free-mode diagnostic run on Qwen2.5-7B-Instruct, which used ALFWorld's built-in \texttt{extra.expert\_plan} API to collect demonstrations. This run is not part of the 12h main leaderboard; it is included only to document a qualitatively different route shape. The run on Qwen3-8B-Base (12h) that achieved 97.76\% did \emph{not} use this API; it relied on model rollout trajectories and prompt format alignment instead (see Appendix~\ref{app:alfworld_trace}). The RL-oriented run (93.28) also did not use this API, as verified by \texttt{grep -r "expert\_plan" code/} returning zero matches.
\begin{alltt}
def collect_expert_demonstrations(env, react_prompts, cfg, num_games):
    rows = []
    for _ in range(num_games):
        ob, info = env.reset()
        ...
        expert_plan = get_expert_plan(info)
        expert_action = expert_plan[0]
        prompt = build_prompt(prefix, initial_observation, history)
        next_ob, _, done, next_info = env.step([expert_action])
        next_expert_plan = get_expert_plan(next_info)
        won = bool(next_info.get("won", [0])[0])
        progress = max(0, len(expert_plan) - len(next_expert_plan))
        weight = 1.0
        weight += 0.05 * max(0, len(expert_plan) - step_idx)
        weight += 0.15 * progress
        weight += 0.80 if won else 0.0
        rows.append({\{}"prompt": prompt, "action": expert_action,
                     "weight": weight, "source": "expert"{\}})
\end{alltt}
\end{promptbox}

\begin{promptbox}[ALFWorld Replay and Weighting Fragment for Qwen2.5-7B-Instruct Free Mode]
\footnotesize
\begin{alltt}
reward += cfg.progress_weight * progress
reward += cfg.success_weight if won else 0.0
reward += cfg.admissible_bonus if admissible else cfg.invalid_penalty
reward += cfg.exact_expert_bonus if action == expert_action else 0.0
reward += cfg.regression_penalty * regression

weight += cfg.success_row_bonus if won else 0.0
weight += cfg.progress_row_bonus * progress
weight += cfg.recovery_row_bonus if trusted_model_action else 0.0
weight += cfg.teacher_row_bonus if acted_by_teacher else 0.0

train_rows = sample_replay_rows(replay_rows, cfg)
run_sft_phase(...)
run_dagger_phase(...)
\end{alltt}
\end{promptbox}

These fragments make clear that the route is not plain SFT. It combines expert warm-starting, online collection under a decaying teacher policy, reward-shaped filtering, and replay-weighted retraining.

\paragraph{WebShop: substantial interactive engineering without gain.}
The representative WebShop failure is also structurally informative. The generated code is not a placeholder; it contains oracle trajectory bootstrap, DAgger-style collection, and weighted behavioral cloning:

\begin{promptbox}[WebShop Interactive Pipeline Fragment]
\footnotesize
\begin{alltt}
oracle_examples, oracle_metrics = collect_oracle_trajectories(
    accelerator=accelerator,
    profiles=profiles[:bootstrap_episodes],
    num_episodes=bootstrap_episodes,
    max_steps=12,
    time_deadline=deadline,
)

dagger_round1, dagger_metrics1 = collect_dagger_examples(
    accelerator=accelerator,
    model=model,
    tokenizer=tokenizer,
    profiles=profiles[: min(512, len(profiles))],
    num_episodes=min(512, len(profiles)),
    max_steps=12,
    model_action_prob=0.30,
    temperature=0.4,
    time_deadline=deadline,
)
\end{alltt}
\end{promptbox}

\begin{promptbox}[WebShop Method Summary Fragment]
\footnotesize
\begin{alltt}
"method": "oracle-trajectory bootstrap + relaxed DAgger with "
          "advantage-weighted behavioral cloning",
"reward_design": {\{}
    "primary": "environment terminal reward from WebShop",
    "state_action_weight": "0.20 + alpha * final_reward * depth_discount "
                           "+ 0.30 * shaped_agreement",
    "bootstrap": "keeps all oracle environment trajectories",
    "dagger": "keeps oracle labels for all visited states and only "
              "adds model actions when the rollout succeeds",
{\}},
\end{alltt}
\end{promptbox}

This negative case is useful because it shows that interactive failure often occurs after substantial engineering effort. The bottleneck is not simply whether code is produced, but whether the resulting reward, data, and training loop produce a stable learning signal.

\paragraph{DeepSearchQA: tool-augmented search trajectory SFT.}
A representative DeepSearchQA route is not PPO-style RL, but it is still structurally informative because it constructs explicit search trajectories rather than plain single-turn answers:

\begin{promptbox}[DeepSearchQA Search-Trajectory Fragment]
\footnotesize
\begin{alltt}
REACT_SYSTEM_PROMPT = """You are a research assistant that
answers complex questions by searching the web.
Thought: ...
Action: search[query]
Action: answer[final answer]"""

def build_search_trajectory(problem: str, answer: str, max_steps: int = 3):
    query = extract_search_query(problem)
    trajectory.append(f"Thought: I need to search ...")
    trajectory.append(f"Action: search[{\{}query{\}}]")
    trajectory.append(f"Observation: {\{}clean_search_result(step1_obs){\}}")
    ...
    trajectory.append(f"Action: answer[{\{}answer{\}}]")
\end{alltt}
\end{promptbox}

\begin{promptbox}[DeepSearchQA SFT Fragment]
\footnotesize
\begin{alltt}
formatted_ds = dataset.map(
    lambda x: {\{}"text": create_prompt(x["problem"], x["answer"]){\}},
    remove_columns=dataset.column_names,
)

trainer = SFTTrainer(
    model=model,
    args=training_args,
    train_dataset=formatted_ds,
)
\end{alltt}
\end{promptbox}

This route is useful in the appendix because it shows that some interactive settings are approached through tool-augmented trajectory construction plus SFT, rather than through explicit policy-gradient RL.

\section{Reproducibility, Assets, and Safeguards}

The benchmark includes several implementation choices intended to make repeated evaluation stable and auditable. Each run is assigned an isolated workspace; submissions are recorded through a grading server rather than by directly overwriting a leaderboard file; and evaluation is serialized within a run to avoid multiple heavyweight evaluators contending for the same GPU simultaneously. The current implementation also caches successful evaluations by resolved path and latest model-file modification time, so that identical submissions are not re-evaluated unnecessarily.

We also explicitly constrain reported evidence to artifacts that can be reconstructed from saved workspaces. Benchmark-native outputs are derived from the grading-server loop, while supplemental process metrics are computed from stored score trajectories and run metadata.

\subsection{Reproducibility and Environment Note}
\label{app:campaign_cost}

The benchmark code, task registry, evaluator implementations, and agent-facing protocol files are maintained in the project repository and are sufficient to reproduce the benchmark logic reported in this paper. The main experimental results additionally depend on infrastructure choices that are not benchmark-specific, including API access to the driver model, local model hosting for the target models, and the cluster environment used to execute long-running agent jobs.

\paragraph{Benchmark-facing environment.}
The benchmark itself assumes a Linux execution environment with Python, PyTorch/Transformers/TRL dependencies, benchmark-specific task packages, and a grading server reachable through a local HTTP endpoint. The main experiments use a 12-hour time budget per run and repeated submissions within that budget. Hardware details matter mainly for runtime and throughput rather than for the definition of the benchmark contract itself.

\paragraph{{Computational cost.}}
{Costs cover the original single-run campaign, excluding repeats and exploratory workspaces. Total GPU-hours include evaluation; training launches may submit multiple checkpoints.}
\begin{center}
\begingroup

\small
\begin{tabular}{@{}lr@{}}
\toprule
Metric & Total \\
\midrule
Run--task cells / allocated run-hours & 78 / 936 h \\
H200-equivalent GPU-hours & 1,404 \\
Driver context tokens & 5.48B \\
Grader submissions & 1,426 \\
Full training launches & 346 \\
Evaluation overhead (GPU-hours) & 102.4 \\
\bottomrule
\end{tabular}
\endgroup
\end{center}

\paragraph{Released artifact manifest.}
The  {public} release is intended to contain the benchmark code, task registry, agent-facing instructions, evaluator and grading-server logic, task configuration files, run scripts, result summaries, selected score traces, and figure-generation scripts. The release does not redistribute upstream benchmark datasets, pretrained base models, internal cluster configuration, or trained model checkpoints. This separation keeps the benchmark contract inspectable while respecting upstream licenses and infrastructure-specific constraints.

\paragraph{What is and is not claimed.}
We treat the benchmark implementation, saved workspaces, and grading artifacts as the reproducible core. We do not make quantitative dollar-cost claims in this version, and we do not claim that cluster orchestration details are necessary to reproduce the benchmark's qualitative conclusions. Where exact infrastructure scripts are environment-specific, we instead document the benchmark contract, the task protocol, and representative run commands in the released codebase.

\paragraph{Existing assets and licenses.}
\benchmarkname{} builds on existing public benchmarks, datasets, environments, and agent frameworks cited in the main text, including GSM8K, HumanEval, AlpacaEval, ALFWorld, WebShop, DeepSearchQA, OpenHands-style agents, OpenCode, and CLI-native agents. The released benchmark code documents concrete dependencies, access paths, and upstream license pointers for each asset; users should follow the original licenses and terms of use. We do not redistribute the original benchmark datasets or pretrained base models as new assets in this paper.

\paragraph{Broader impact.}
The intended positive impact of \benchmarkname{} is to make agent-driven post-training more transparent and auditable by exposing not only final model scores but also failures in planning, debugging, rollout collection, and reward handling. A possible negative impact is that better diagnostics may accelerate more capable autonomous post-training agents, including systems that could be misused if paired with unsafe objectives. We mitigate this by releasing an evaluation framework and audit artifacts rather than trained high-risk models, and by emphasizing limitations, reproducibility, and failure analysis.

\subsection{Audit and Anti-Cheating Rules}
\label{app:data_integrity}

Because \benchmarkname{} evaluates autonomous agents with repeated submissions, the integrity rules need to be explicit. Table~\ref{tab:audit_rules} summarizes the safeguards that are currently implemented and auditable in the released system.

\begin{table*}[t]
\centering
\caption{Implemented audit and anti-cheating rules.}
\label{tab:audit_rules}
\small
\setlength{\tabcolsep}{4pt}
\begin{tabular}{@{}P{0.22\textwidth}P{0.72\textwidth}@{}}
\toprule
Rule category & Implemented behavior \\
\midrule
Test-set isolation & Held-out evaluation data are not mounted into the agent workspace. Agents only receive training-visible data and task descriptions. \\
Submission interface & Scores are produced only through the grading server. Agents cannot directly overwrite leaderboard files or final result tables. \\
Baseline reuse & Trivial submission of the untouched base model is treated as non-improving behavior and maps back to baseline performance. \\
Artifact logging & Each run stores score trajectories, run metadata, and workspace artifacts for later inspection. \\
Evaluator serialization & Heavyweight evaluation is serialized within a run to reduce contention and make score traces easier to audit. \\
Path-based caching & Identical submissions may be cached by resolved path and file-modification state to avoid repeated redundant evaluation, without changing the best-score semantics. \\
\bottomrule
\end{tabular}
\end{table*}

These safeguards are intended to reduce straightforward leakage or score inflation, not to claim perfect adversarial security. We therefore restrict the main paper's integrity claims to mechanisms that are already present in the implementation and can be inspected from the released code and artifacts.

\subsection{Failure Taxonomy Summary}
\label{app:failure_taxonomy}

The main paper argues that the interactive boundary is exposed not by a single failure mode, but by several recurrent breakdowns in the online loop. Table~\ref{tab:failure_taxonomy} summarizes the dominant categories we repeatedly observed when auditing saved workspaces and score traces.

\begin{table*}[t]
\centering
\caption{Failure taxonomy used for offline diagnosis of interactive-task runs.}
\label{tab:failure_taxonomy}
\small
\setlength{\tabcolsep}{3pt}
\begin{tabular}{@{}P{0.18\textwidth}P{0.15\textwidth}P{0.31\textwidth}P{0.29\textwidth}@{}}
\toprule
Failure type & Typical task & Observable symptom & Main audit artifact \\
\midrule
Rollout construction failure & ALFWorld, WebShop & The run produces training code, but the environment loop is incomplete, brittle, or never yields a usable trajectory set. & \texttt{train.py}, task-specific \texttt{eval.py}, missing or degenerate submissions \\
Reward/data design failure & WebShop, DeepSearchQA & The run implements non-trivial collection and training logic, yet the resulting data or reward signal is too sparse, brittle, or misaligned to improve performance. & Generated training script, score trajectory, failure notes in \texttt{summary.md} \\
Training collapse / instability & WebShop, ALFWorld & Early submissions exist, but later iterations regress or fail during model loading, runtime startup, or evaluation. & \texttt{scores.json}, run logs, final workspace artifacts \\
Early coding failure & All tasks, especially weaker-model settings & The run terminates before producing substantive training code or a valid submission. & Missing or placeholder \texttt{train.py}, zero-submission runs \\
Static shortcut without transfer & GSM8K, HumanEval versus L3 tasks & The scaffold can assemble a workable static SFT/RL route, but the same competence does not carry over once interactive rollout becomes mandatory. & Cross-task comparison of route labels, submissions, and improvements \\
\bottomrule
\end{tabular}
\end{table*}

This taxonomy is intentionally diagnostic rather than benchmark-native. It is used to interpret why interactive tasks behave differently from static ones, not to redefine the benchmark score itself.

\section{Exploratory Agent Behavior Diagnostics}
\label{app:cc_behavior}

Table~\ref{tab:agent_behavior} summarizes exploratory Claude Code workspaces used for qualitative behavior analysis. These numbers are derived from saved artifacts (\texttt{scores.json}, \texttt{code/}, \texttt{output/}) and are not part of the 12h main leaderboard or scaffold comparison.

\begin{table*}[t]
\centering
\caption{Exploratory Claude Code workspace statistics per task. Submissions = total valid grading requests; Code files = distinct \texttt{.py} files written under \texttt{code/}; Model versions = directories under \texttt{output/}; Zero rate = fraction of submissions scoring 0 because training or evaluation did not complete successfully; Best@Sub = submission number at which the best score was achieved.}
\label{tab:agent_behavior}
\small
\setlength{\tabcolsep}{3pt}
\begin{tabular}{@{}lccccccP{0.25\textwidth}@{}}
\toprule
Task & Mode & Subs & Code & Models & Zero\% & Best@Sub & Primary Strategies \\
\midrule
GSM8K & SFT & 92 & 14 & 146 & 20\% & \#90 & SFT, model soup, seed sweep \\
HumanEval & SFT & 195 & 28 & 336 & 23\% & \#26 & SFT, model averaging, task arithmetic \\
AlpacaEval & Free & 23 & 11 & 47 & 0\% & \#23 & SFT $\to$ DPO, synthetic preference gen. \\
ALFWorld & Free & 4 & 4 & 5 & 0\% & \#4 & Demonstration SFT (env.\ interaction) \\
WebShop & Free & 16 & 4 & 20 & 19\% & \#13 & Trajectory collection $\to$ SFT \\
DeepSearchQA & Free & 145 & 47 & 285 & 0\% & \#39 & DPO, GRPO, KTO, SLERP, RFT \\
\midrule
\textbf{Total} & & \textbf{475} & \textbf{108} & \textbf{839} & \textbf{{14\%}} & & \\
\bottomrule
\end{tabular}
\end{table*}

\paragraph{Exploration volume varies by orders of magnitude.}
The most compact success (ALFWorld: 4 submissions, 4 code files) and the most prolific exploration (HumanEval: 195 submissions, 336 model versions) differ by $\sim$50$\times$ in submission count. This range reflects genuinely different task structures: ALFWorld's demonstration learning provides a direct path, while HumanEval and DeepSearchQA require search over training configurations.

\paragraph{Late peaking vs.\ early peaking.}
Best scores arrive at qualitatively different stages. GSM8K peaks at submission \#90/92 (98th percentile, near the end), while HumanEval peaks at \#26/195 (13th percentile, early). ALFWorld peaks at \#4/4 (monotonic), and DeepSearchQA at \#39/145 (27th percentile). Late-peaking tasks (GSM8K) suggest that patient hyperparameter search pays off; early-peaking tasks (HumanEval) suggest diminishing returns despite continued exploration.

\paragraph{Zero-score rate as a fragility indicator.}
HumanEval (23\%) and GSM8K (20\%) have the highest zero-score rates, caused primarily by OOM failures during training. The agent recovers from these consistently by changing batch size, gradient checkpointing, or adapter-based training. By contrast, AlpacaEval and DeepSearchQA achieve zero such failures across 23 and 145 submissions respectively, indicating that the agent successfully manages resource constraints for these tasks.

\subsection{Claude Code Per-Task Case Studies}
\label{app:cc_cases}

We describe the representative agent trajectory for each task, drawing on workspace artifacts. These cases are diagnostic: they complement the quantitative results in the main text but are not benchmark-native metrics.

\paragraph{Case 1: GSM8K, model soup over 92 submissions.}
The representative SFT-mode run produced 146 model versions across 14 code files. Its strategy evolved from basic SFT $\to$ multi-seed SFT $\to$ weighted model averaging (model soup). The agent wrote averaging scripts that combined 2 to 4 checkpoints with learned interpolation weights. The best score (83.85) was reached at submission \#90, following a model soup of three checkpoints trained with different learning rates and seeds. The 20\% zero-score rate was caused by OOM failures on larger batch sizes; the agent systematically reduced batch size and enabled gradient checkpointing after each failure. In a representative Free-mode run, the same task was approached via GRPO with 15 distinct reward function variants, but only reached 82.34, suggesting that the additional complexity of reward engineering does not pay off when the task's difficulty is primarily in data construction.

\paragraph{Case 2: HumanEval, most prolific exploration.}
The representative SFT-mode run was the most prolific across all experiments: 195 submissions, 336 output model versions, 28 code files. The agent's code evolution reveals a progression: basic SFT $\to$ data augmentation (generating additional problems) $\to$ model averaging $\to$ task arithmetic (combining delta vectors from different training runs) $\to$ seed sweeps. The 45 zero-score failures (23\%) were predominantly OOM errors; the agent demonstrated consistent recovery by reducing batch size, adding gradient checkpointing, or switching to LoRA. Despite this prolific exploration, the best score (81.71) was reached at submission \#26, with the subsequent 169 submissions unable to improve, a textbook optimization plateau where additional computation is wasted.

\paragraph{Case 3: AlpacaEval, synthetic preference data breakthrough.}
The representative Free-mode run exhibited the most controlled trajectory: 23 submissions, zero zero-score failures, monotonically improving score trajectory from 20.01 to 31.02. The key innovation was synthetic preference data generation. The agent wrote \texttt{gen\_claude\_data.py} and \texttt{gen\_gpt4style\_data.py} to generate diverse candidate responses to AlpacaEval instructions, then used an LLM judge to create pairwise preference annotations. This data was used for DPO training after an initial SFT warm-up. The SFT-only mode, lacking DPO, explored 102 submissions with alternative strategies (NEFTune perturbation, SLERP model merging, task arithmetic) across 130 model versions but only reached 21.10, a 10-point gap that directly quantifies the value of the composite SFT$\to$DPO pipeline.

\paragraph{Case 4: ALFWorld, demonstration learning via environment interaction.}
The representative Free-mode run collected environment demonstrations through interaction and rollout, and achieved 95.52 in just 4 submissions. In RL-oriented mode, the agent followed a two-phase SFT$\to$GRPO pipeline, reaching 93.28. In SFT-only mode, the agent managed 67.91, suggesting that the critical distribution alignment insight is not deterministic; even the same scaffold may or may not find it depending on the exploration path.

\paragraph{RL-oriented ALFWorld workspace audit.}
\label{app:alfworld_audit}
Because the RL-oriented ALFWorld result (93.28) serves as the paper's primary evidence that an agent can engineer an online RL loop, we provide an explicit workspace audit. The audited RL-oriented workspace produced multiple candidate submissions and generated dedicated rollout-collection and GRPO training code:
\begin{itemize}
\item \textbf{No privileged API usage}: audit greps over generated code for privileged or oracle-like ALFWorld shortcuts, including \texttt{expert\_plan}, \texttt{admissible\_commands}, \texttt{extra.gamefile}, \texttt{oracle}, and \texttt{gold\_trajectory}. These checks returned no usage in the RL-oriented run. The agent collected demonstrations through environment interaction (\texttt{collect\_demos.py}) using the model's own rollouts.
\item \textbf{SFT warm-up}: Initial SFT on self-collected demonstrations, producing a baseline for further improvement.
\item \textbf{GRPO with online rollouts}: The agent implemented \texttt{grpo\_train.py} using \texttt{GRPOTrainer} from TRL with reward signals derived from environment success/failure. Online rollout collection was performed between GRPO iterations.
\item \textbf{Trajectory-level rewards}: The reward function used sparse binary environment feedback (task success = +1, failure = 0), which is the canonical RL setup for ALFWorld.
\item \textbf{Progressive improvement}: The score trajectory improved from the 4.85 baseline through strong SFT and GRPO variants, with representative milestones reaching 88.81, 90.30, 91.79, and finally 93.28. Later diagnostic variants did not improve the best score.
\end{itemize}
This audit confirms that the RL-oriented result represents genuine online RL engineering, including environment interaction, trajectory collection, and reward-driven optimization. In this paper, RL-oriented means that the agent is instructed to use RL or online reward optimization as the primary improvement mechanism, while SFT or behavior-cloning warm-up is allowed when needed to make the RL loop stable. The constraint was enforced through prompt instruction rather than API removal, so this is an audited compliance result rather than a benchmark-enforced guarantee.

\paragraph{Case 5: WebShop, sign reversal through trajectory quality.}
The representative Free-mode run achieved 84.0 in 16 submissions with a trajectory-collection-then-SFT pipeline. The agent wrote dedicated scripts for trajectory collection (\texttt{collect\_data.py}), quality filtering (\texttt{process\_data.py}), and SFT training. Three zero-score submissions (19\%) occurred when evaluation did not complete within the run constraints. The SFT-mode agent achieved 76.0 after collecting and filtering trajectories, while the Free-mode route remained the strongest 7B WebShop result. This sign reversal from OpenCode-7B suggests that the primary bottleneck is trajectory collection quality, not training algorithm choice.

\paragraph{Case 6: DeepSearchQA, exhaustive technique search.}
The representative Free-mode run was the most technically diverse: 47 code files implementing DPO (v1 to v16), GRPO, KTO, ORPO, rejection fine-tuning, NEFTune, Adafactor, 3-way SLERP merging, 4-way TIES merging, and task arithmetic. Despite producing 285 model versions and 145 submissions, the best 7B-mode score reached only 15.0 ($+3.25$). A representative RL-oriented run independently converged to the same score via 117 submissions with DPO/GRPO/SFT/KTO pipelines. This convergent difficulty across diverse strategies, two independent agents, and two modes suggests that DeepSearchQA's search-retrieve-judge loop resists improvement through post-training alone, likely requiring architectural or inference-time interventions beyond what current agent-driven post-training can provide.

\paragraph{DeepSearchQA follow-up: SFT sensitivity and low absolute scores.}
\label{app:dsqa_v12}
A controlled follow-up experiment pits Claude Code (Opus 4.6 driver) against Codex (GPT-5.4 driver) on DeepSearchQA with Qwen3-8B-Base (baseline=12.25), each given 12 hours on 2$\times$H200. Updated endpoints reach 23.0 for Claude Code and 21.0 for Codex, improving over early local sweeps but still remaining low relative to the other interactive tasks.

\begin{table}[h]
\centering
\caption{DeepSearchQA SFT sensitivity analysis from early local sweeps (Qwen3-8B-Base, 12h). The optimization landscape has a narrow sweet spot; deviations in any direction degrade performance sharply.}
\label{tab:dsqa_sweetspot}
\small
\begin{tabular}{lccc}
\toprule
Parameter & Under-fit & Sweet spot & Over-fit \\
\midrule
Learning rate & 5e-5 $\to$ 10.5 & \textbf{1e-4 $\to$ 11 to 13} & 2e-4 $\to$ 4.0 \\
LoRA rank & 16 $\to$ 10.5 & \textbf{32 $\to$ 11 to 13} & 64 $\to$ 4.0 to 7.5 \\
Epochs & 2 $\to$ 10.5 & \textbf{3 to 4 $\to$ 11 to 13} & 5+ $\to$ 6.5 \\
Training loss & $>$1.5 & \textbf{1.1 to 1.2} & $<$0.5 \\
\bottomrule
\end{tabular}
\end{table}

The key diagnostic finding is that DeepSearchQA has a narrow SFT sweet spot and does not consistently reward more complex post-training routes in this setting. Early Codex sweeps clustered around 11 to 13 points, while deviations in learning rate, LoRA rank, epoch count, or training loss degraded sharply. Later endpoints improve to 21 to 23, but the task remains lower-scoring than ALFWorld and WebShop. The reward signal from the LLM judge is sparse: many completions receive similar scores, producing weak gradients for online RL methods on small models.

\section{Additional Quantitative Results}

\subsection{Claude Code Mode Sensitivity}
\label{app:cc_mode_sensitivity}

\begin{figure}[t]
\centering
\includegraphics[width=0.98\linewidth]{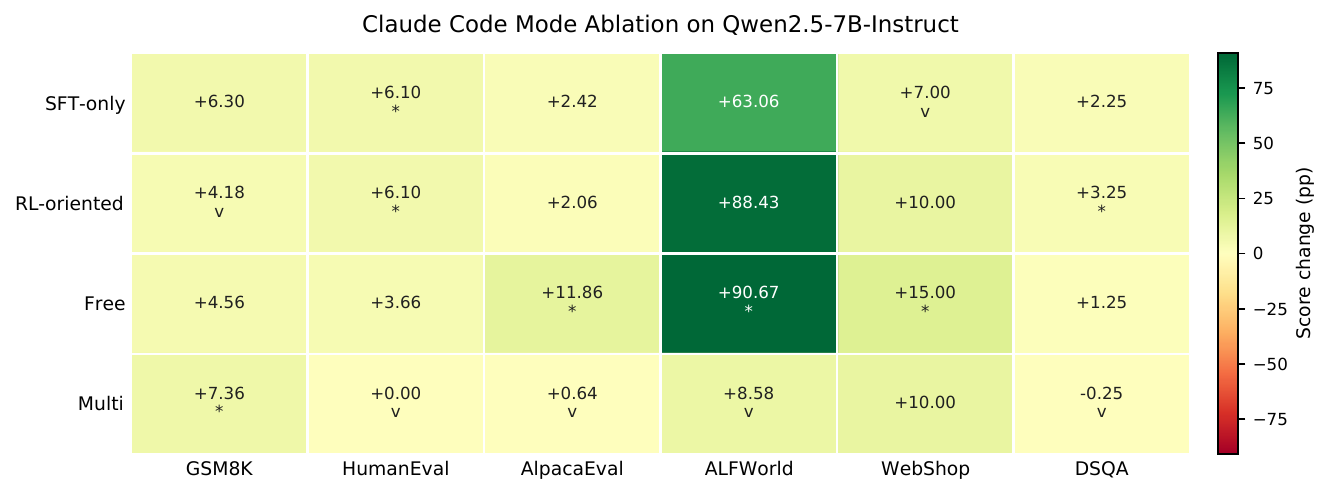}
\caption{Mode $\times$ Task improvement heatmap for Claude Code. Stars mark the best mode per task; triangles mark the worst. Rank reversals between modes confirm that no single training paradigm uniformly dominates.}
\label{fig:mode_heatmap}
\end{figure}

Figure~\ref{fig:mode_heatmap} provides a compact view of the operating-mode reversal reported in Table~\ref{tab:mode_comparison}. Static tasks have modest spreads across modes, while ALFWorld and WebShop separate modes sharply. This supports the main claim that agent-driven post-training results should report the allowed operating mode, since the same scaffold and driver can look substantially different under SFT-only, RL-oriented, Free, and Multi-task constraints.

\subsection{Additional 12h Experimental Results}
\label{app:additional_results}

The following tables and analyses  {report single-run results under a 12h budget}. They provide supporting evidence for the main claims without changing the primary leaderboard.

\begin{table}[h]
\centering
\caption{RL method audit across Claude Code workspaces. ``Stable'' means that training completed reliably and produced non-degenerate results.}
\label{tab:rl_audit}
\small
\setlength{\tabcolsep}{3pt}
\begin{tabular}{lP{3.5cm}cP{4.3cm}}
\toprule
Task & RL Methods Tried & Stable? & Best Pipeline \\
\midrule
GSM8K & GRPO ($\times$15 variants) & Partial & SFT + model soup \\
HumanEval & GRPO, DPO, RFT & No & SFT + model averaging \\
AlpacaEval & GRPO, DPO ($\times$31) & Yes (DPO) & SFT $\to$ DPO \\
ALFWorld & GRPO + online rollout & Yes & SFT $\to$ GRPO (RL-oriented) \\
WebShop & DPO, ORPO, KTO & Partial & Traj.\ collection $\to$ SFT \\
DeepSearchQA & DPO, GRPO, KTO, ORPO & Partial & SFT $\to$ DPO pipeline \\
\bottomrule
\end{tabular}
\end{table}

\paragraph{{Gemini driver comparison.}}
{We compare Gemini 2.5 Flash and Gemini 3.1 Pro Preview within Gemini CLI, fixing Qwen3-8B-Base, the six tasks, grader, submission interface, checkpoint-selection rule, and 12h budget.}

\begin{table}[ht]
\centering
\caption{{Single-run Gemini CLI comparison on Qwen3-8B-Base (12h). Avg $\Delta$ is the unweighted mean improvement over the base-model baseline across six tasks (pp).}}
\label{tab:gemini_driver_comparison}
\begingroup

\small
\setlength{\tabcolsep}{3pt}
\resizebox{\textwidth}{!}{
\begin{tabular}{lrrrrrrr}
\toprule
Driver & GSM8K & HumanEval & AlpacaEval & ALFWorld & WebShop & DeepSearchQA & Avg $\Delta$ \\
\midrule
Gemini 2.5 Flash & 70.13 & 10.98 & 7.14 & 11.19 & 0.00 & 13.00 & $-11.45$ \\
Gemini 3.1 Pro Preview & 87.49 & 79.27 & 23.08 & 86.57 & 68.00 & 20.50 & $+30.63$ \\
\bottomrule
\end{tabular}}
\endgroup
\end{table}

\begin{table}[h]
\centering
\caption{Agent comparison on DeepSearchQA (Qwen3-8B-Base, 12h).}
\label{tab:agent_efficiency}
\small
\setlength{\tabcolsep}{4pt}
\begin{tabular}{lcc}
\toprule
Dimension & Claude Code & Codex \\
\midrule
Driver LLM & Claude Opus 4.6 & GPT-5.4 \\
Best $\Delta$ & +10.75 & +8.75 \\
\bottomrule
\end{tabular}
\end{table}

\paragraph{Scaling diagnostics.}
Most gains on static tasks arrive within the first six hours, while interactive tasks more often depend on later progress within the fixed 12h budget (Figure~\ref{fig:combined_scaling} in the main text). The scaling figures show performance as a function of time, context-token counter, and submission count, enabling multi-dimensional efficiency comparison across systems. On ALFWorld, Codex+GPT-5.4 reaches {91.04} and Codex+GPT-5.2 reaches 87.31, while CC+Sonnet~4.5 remains near baseline despite substantial process budget. On static tasks (GSM8K, HumanEval), most systems plateau early in the run.

\begin{figure*}[t]
\centering
\includegraphics[width=0.98\textwidth]{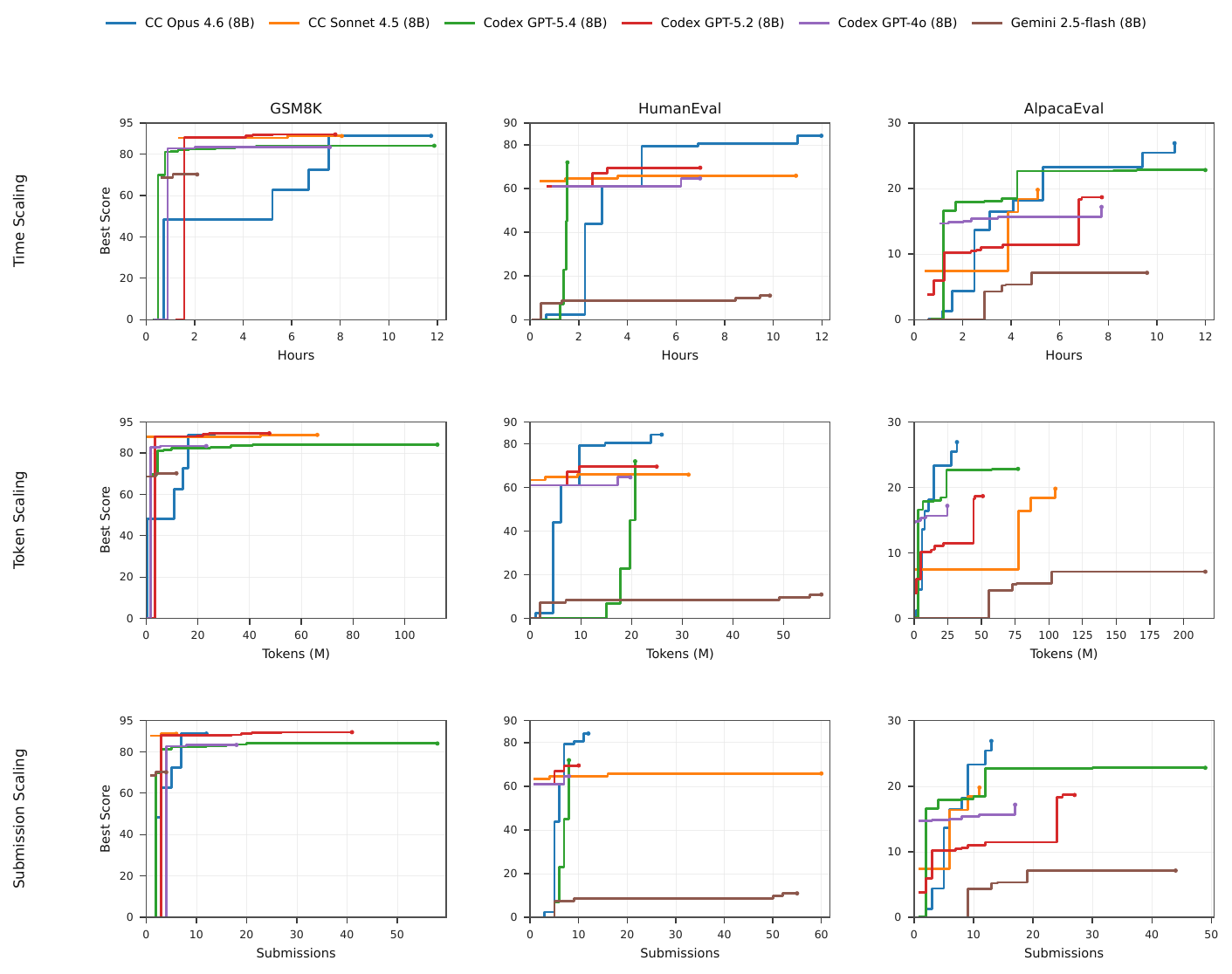}
\caption{Scaling analysis for the three static tasks across all agent stacks. \textit{Top:} Time scaling. \textit{Middle:} Token scaling. \textit{Bottom:} Submission scaling. Compared with the interactive-task scaling in Figure~\ref{fig:combined_scaling}, static tasks typically plateau earlier and show weaker separation between strong agent stacks.}
\label{fig:static_scaling}
\end{figure*}

\begin{figure}[t]
\centering
\includegraphics[width=0.9\linewidth]{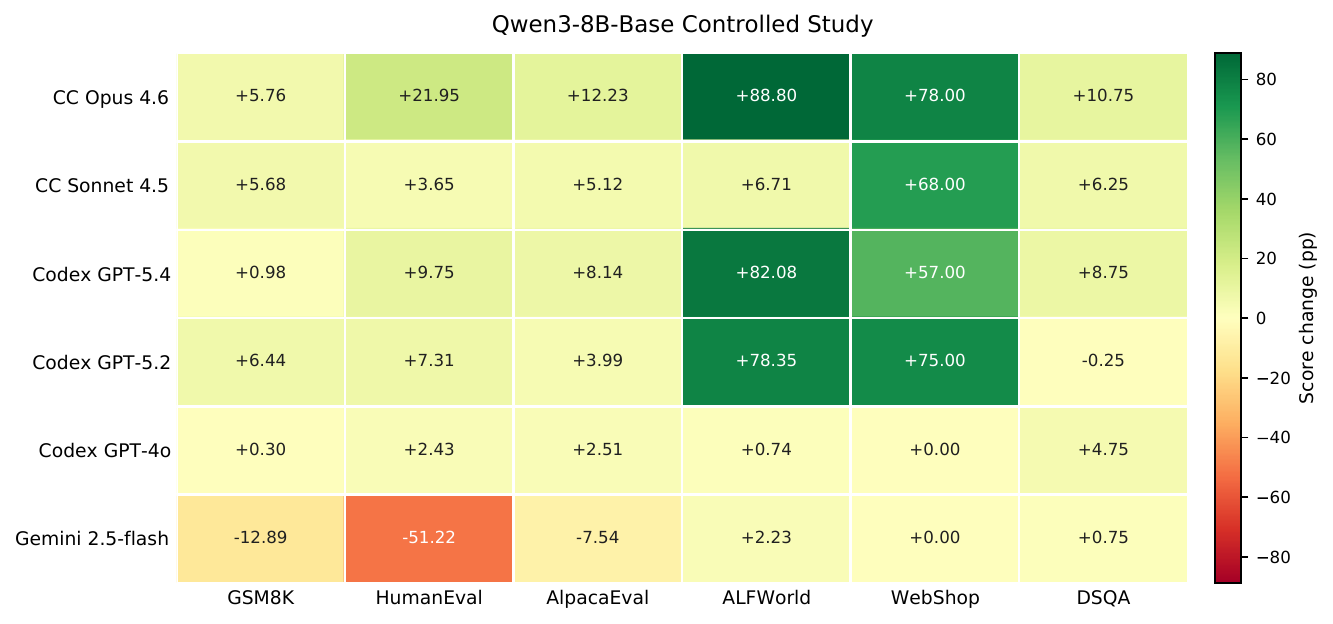}
\caption{Improvement over canonical Qwen3-8B-Base baseline ($\Delta$ in percentage points) for six agent stacks across all tasks (12h). {The CC Opus GSM8K result and Table~\ref{tab:mode_comparison}'s Free-mode result come from separate runs.} Interactive tasks (ALFWorld, WebShop) show larger observed differences across driver and scaffold choices than static tasks.}
\label{fig:scaffold_driver_heatmap}
\end{figure}

\subsection{Method Attribution and Route-Type Diagnostic}
\label{app:route_diagnostic}

We complement the score-level leaderboard with a route-type diagnostic. Every submission is annotated with the training method recovered from the agent's own workspace \texttt{git} history (\texttt{exp\_NN} commit messages describe the training approach used for each submission). Methods are classified into eight categories: SFT (plain), Model merging (SLERP / TIES / soup / weight interpolation), RFT (rejection fine-tuning), GRPO, DPO, KTO / ORPO, PPO, and Other. This annotation underpins the main-text observation that {in the original main campaign, }only ALFWorld RL-oriented contains a strict online-RL component among the best-scoring routes.

\begin{table}[h]
\centering
\caption{Method attempt vs.\ adoption frequency across reconstructable Claude Code 7B cells. "Attempted" counts cells in which at least one submission is tagged with the method; "Adopted in best" counts cells whose highest-scoring submission used the method.}
\label{tab:method_adoption}
\small
\begin{tabular}{lccc}
\toprule
Method & Attempted cells & Adopted in best & Adoption rate \\
\midrule
SFT (plain) & 19 & 18 & 95\% \\
Model merging & 11 & 4 & 36\% \\
RFT & 5 & 3 & 60\% \\
GRPO & 8 & 2 & 25\% \\
DPO & 4 & 1 & 25\% \\
KTO / ORPO & 3 & 0 & 0\% \\
PPO & 1 & 0 & 0\% \\
Other & 2 & 0 & 0\% \\
\bottomrule
\end{tabular}
\end{table}

Across reconstructable 7B Claude Opus cells, every best-scoring route contains SFT as the sole mechanism or as an SFT-initialized warm start. Online-RL methods (GRPO / PPO) are attempted in roughly nine of twenty-four cells but adopted in only two, and in both cases they require SFT warm-up. This is the empirical basis for the main-text claim that score improvement and online-RL engineering are separable capabilities.

Figure~\ref{fig:trajectory_methods} visualizes the per-submission trajectory alongside method-switch events for three representative tasks. Stepwise score jumps coincide with cross-module engineering interventions (prompt-format alignment, OOM-driven LoRA adoption, route pivots) rather than with smooth hill-climbing against grader feedback, which we take as qualitative evidence that reported gains reflect agent engineering rather than adaptive overfitting to the submission API.

\clearpage
\subsection{{Matched Trajectory-SFT and Online RL}}
\label{app:matched_online_rl}

{We compare trajectory-SFT and audited online GRPO using Claude Code with Opus 4.6 on Qwen2.5-7B-Instruct. Each pair branches from the same warm-up data and checkpoint, fixing the 12h budget, GPU configuration, environment-interaction allowance, training-token budget, submission cap, and checkpoint-selection rule. The online branch includes target-policy environment rollouts, episode-level rewards, and online policy updates.}

\begin{table}[!ht]
\centering

\caption{{Matched trajectory-SFT versus audited online RL. Scores: mean $\pm$ sample SD over three paired runs. Gain: mean paired improvement (pp).}}
\label{tab:matched_online_rl}
\small
\setlength{\tabcolsep}{5pt}
\begin{tabular}{lcccc}
\toprule
Task & Baseline & Trajectory-SFT & Online RL & Gain (pp) \\
\midrule
ALFWorld & 4.85 & $88.06 \pm 1.97$ & $92.29 \pm 2.40$ & $+4.23$ \\
WebShop & 69.00 & $77.67 \pm 1.53$ & $81.67 \pm 1.53$ & $+4.00$ \\
\bottomrule
\end{tabular}
\end{table}

Four independent ALFWorld online-RL runs achieved $91.07 \pm 2.44\%$ (mean $\pm$ sample SD) when their frozen checkpoints were each evaluated once on 140 \texttt{valid\_seen} episodes withheld from training and checkpoint selection.

\subsection{Run-to-Run Variability}
\label{app:run_variability}

\begin{table}[!ht]
\centering
\caption{Sample SDs over four runs for Table~\ref{tab:main_results}. Avg $\Delta$ SD is computed across run-level six-task averages.}
\label{tab:main_results_sd}
\small
\setlength{\tabcolsep}{3.5pt}
\resizebox{\textwidth}{!}{%
\begin{tabular}{llccccccc}
\toprule
System & Driver & GSM8K & HumanEval & AlpacaEval & ALFWorld & WebShop & DeepSearchQA & Avg $\Delta$ \\
\midrule
\multicolumn{9}{l}{\textit{Panel A: Qwen2.5-7B-Instruct}} \\
\midrule
OpenHands & GPT-5.4 & 0.78 & 2.08 & 1.03 & 2.15 & 5.35 & 1.03 & 0.78 \\
OpenCode & GPT-5.4 & 1.75 & 1.41 & 1.57 & 3.97 & 5.62 & 1.80 & 1.51 \\
Claude Code & Opus 4.6 & 1.56 & 3.64 & 2.43 & 2.24 & 5.38 & 1.11 & 0.99 \\
\midrule
\multicolumn{9}{l}{\textit{Panel B: Qwen3-8B-Base}} \\
\midrule
Claude Code & Opus 4.6 & 1.86 & 3.79 & 2.35 & 3.31 & 5.50 & 2.32 & 0.15 \\
Claude Code & Sonnet 4.5 & 1.32 & 0.61 & 0.34 & 37.25 & 7.85 & 4.09 & 6.56 \\
Codex CLI & GPT-5.4 & 1.56 & 2.08 & 2.12 & 3.02 & 3.59 & 2.35 & 1.14 \\
Codex CLI & GPT-5.2 & 1.38 & 5.50 & 1.52 & 5.70 & 4.08 & 2.25 & 1.78 \\
Codex CLI & GPT-4o & 1.19 & 5.89 & 0.63 & 3.19 & 17.69 & 0.96 & 2.75 \\
Gemini CLI & 2.5-flash & 4.86 & 6.32 & 2.18 & 1.65 & 12.96 & 1.68 & 2.22 \\
\bottomrule
\end{tabular}}
\end{table}

\section{Detailed Trace Walkthroughs}

\subsection{HumanEval on Qwen3-8B-Base}
\label{app:humaneval_trace}

The per-task case studies above summarize behavioral patterns at the workspace level. To complement this, we present a fine-grained walkthrough of a single run that illustrates both the strengths and limitations of LLM-agent-driven post-training in concrete detail.

\paragraph{Setup.}
The trace comes from a Claude Code Free-mode run on HumanEval with Qwen3-8B-Base, an unaligned base model that scores 62.20\% pass@1 under the held-out 82-problem evaluation split. The agent was given a 12-hour budget on a single H200 GPU. Table~\ref{tab:trace_summary} summarizes the progression, including early submissions that regressed below the base model before later recovery.

\begin{table}[h]
\centering
\caption{Submission progression for the HumanEval trace walkthrough (Claude Code Free, Qwen3-8B-Base). Each row corresponds to a qualitatively distinct strategy shift.}
\label{tab:trace_summary}
\begin{tabular}{clcl}
\toprule
Sub \# & Script & Score & Method \\
\midrule
baseline & N/A & 62.20\% & Qwen3-8B-Base (unaligned) \\
\#3 & \texttt{sft\_train.py} & 2.44\% & SFT v1 \\
\#5 & \texttt{sft\_train\_v3.py} & 43.90\% & SFT v3 (prompt alignment) \\
\#7 & \texttt{grpo\_train.py} & 79.27\% & GRPO on SFT checkpoint (LoRA) \\
\#11 & \texttt{rft\_train.py} & \textbf{84.15\%} & RFT (rejection sampling fine-tuning) \\
\bottomrule
\end{tabular}
\end{table}

\paragraph{Step 1: SFT v1 (Sub \#3, 2.44\%).}
The agent's first move was supervised fine-tuning on the 82 training-visible HumanEval examples, teaching the base model to produce a \texttt{<think>} tag followed by code:

\begin{promptbox}[SFT v1 Training Format]
\footnotesize
\begin{alltt}
messages = [
    \{"role": "user",
     "content": f"Complete the following Python function:\textbackslash{}n\textbackslash{}n\{question\}"\},
    \{"role": "assistant",
     "content": f"<think>\textbackslash{}nI need to implement...\textbackslash{}n</think>\textbackslash{}n\textbackslash{}n\{answer\}"\}
]
\end{alltt}
\end{promptbox}

Training used 4 epochs, lr=$2 \times 10^{-5}$, batch size 2, gradient accumulation 4. The model learned to generate output, but the submitted checkpoint scored only 2.44\% because the evaluator could not parse most responses, since the training prompt format did not match the evaluation prompt format.

\paragraph{Step 2: SFT v3 with prompt alignment (Sub \#5, 43.9\%).}
The agent diagnosed the format mismatch and aligned the training prompt to match the evaluation prompt exactly:

\begin{promptbox}[Prompt Alignment Change]
\footnotesize
\begin{alltt}
# Before (mismatched with evaluator)
"Complete the following Python function:\textbackslash{}n\textbackslash{}n\{question\}"

# After (aligned with evaluator)
EVAL_PROMPT_TEMPLATE = (
    "Read the following function signature and docstring, "
    "and fully implement the function described. Your response "
    "should only contain the code for this function.\textbackslash{}n"
    "\{question\}"
)
\end{alltt}
\end{promptbox}

This single string change {accompanied} an 18$\times$ improvement over the previous trained submission (2.44\% $\to$ 43.9\%). The agent also applied data augmentation, expanding 82 examples to 246 by generating three output format variants per example:

\begin{promptbox}[Data Augmentation Formats]
\footnotesize
\begin{alltt}
# Format 1: complete function (with signature)
f"<think>\textbackslash{}n\{entry_point\}\textbackslash{}n</think>\textbackslash{}n\textbackslash{}n```python\textbackslash{}n\{question\}\{answer\}```"
# Format 2: function body only
f"<think>\textbackslash{}nImplement \{entry_point\}.\textbackslash{}n</think>\textbackslash{}n\textbackslash{}n\{answer\}"
# Format 3: minimal thinking
f"<think>\textbackslash{}nOK\textbackslash{}n</think>\textbackslash{}n\textbackslash{}n\{answer\}"
\end{alltt}
\end{promptbox}

\textbf{Agent limitation:} This augmentation introduces \emph{training target inconsistency}. Format~1 trains the model to output complete functions (including the signature), while Formats~2 and 3 train it to output only the function body, creating two conflicting objectives in the same training set. The \texttt{<think>} content is a placeholder (the function name), not genuine reasoning. The {41.46pp} improvement {may be partly} attributable to prompt alignment, but the agent does not perform this ablation.

The downstream GRPO reward function later compensates for the mixed output format via post-processing:

\begin{promptbox}[Post-Processing Workaround for Mixed Formats]
\footnotesize
\begin{alltt}
if f"def \{meta['entry_point']\}" in code_body:
    full_function = code_body           # complete function
else:
    full_function = meta["question"] + code_body  # body only
\end{alltt}
\end{promptbox}

This workaround masks the training inconsistency rather than resolving it at the source, a pattern we observe across multiple agent runs where post-processing substitutes for clean data design.

\paragraph{Step 3: GRPO with code execution reward (Sub \#7, 79.27\%).}
With a 43.9\% SFT checkpoint as the starting point, the agent introduced reinforcement learning via GRPO using actual code execution as the reward signal:

\begin{promptbox}[GRPO Reward Function]
\footnotesize
\begin{alltt}
def reward_function(completions, **kwargs):
    code_body = extract_code(assistant_text)
    program = (full_function + "\textbackslash{}n\textbackslash{}n" + meta["test"]
               + f"\textbackslash{}n\textbackslash{}ncheck(\{meta['entry_point']\})\textbackslash{}n")
    passed = _run_code(program, timeout=10)
    rewards.append(1.0 if passed else 0.0)
\end{alltt}
\end{promptbox}

\textbf{Agent strength: memory management.}
{The run used }a single H200 ({141}\,GB). The agent switched to LoRA ($r{=}16$) {to reduce training memory requirements}:

\begin{promptbox}[LoRA Configuration for GRPO]
\footnotesize
\begin{alltt}
lora_config = LoraConfig(
    r=16, lora_alpha=32,
    target_modules=["q_proj","k_proj","v_proj","o_proj",
                    "gate_proj","up_proj","down_proj"],
)
config = GRPOConfig(
    num_generations=4,        # 4 samples per prompt
    temperature=0.8,
    max_completion_length=768,
    num_train_epochs=3,
    learning_rate=5e-5,
)
\end{alltt}
\end{promptbox}

After training, the agent merged LoRA weights back into the base model before submission. The result: 79.27\%, a 35-point gain over SFT alone.

\paragraph{Step 4: RFT after GRPO degradation (Sub \#11, 84.15\%).}
Submissions \#8 to \#10 continued GRPO iterations but showed degradation (down to 76.83\%), indicating the policy had reached a local optimum. The agent recognized this and switched to rejection fine-tuning (RFT): sampling 16 candidate solutions per problem from the best checkpoint (Sub \#{9}, 80.49\%), filtering by test execution, and fine-tuning on the verified solutions.

\begin{promptbox}[RFT Sampling and Filtering]
\footnotesize
\begin{alltt}
NUM_SAMPLES = 16
for ex in raw_data:
    for batch_start in range(0, NUM_SAMPLES, 4):
        outputs = model.generate(
            **inputs, max_new_tokens=768, temperature=0.8,
            top_p=0.95, do_sample=True, num_return_sequences=4,
        )
        for output in outputs:
            code_body = extract_code(output_text)
            if verify_solution(code_body, meta):
                correct_solutions.add(output_text)
\end{alltt}
\end{promptbox}

This yielded 993 verified model-generated solutions plus 82 ground-truth examples (1,075 total). The agent fine-tuned with a conservative learning rate ($1 \times 10^{-6}$, lower than the $5 \times 10^{-6}$ used in SFT v3) to avoid overwriting GRPO-learned capabilities, reaching the final score of 84.15\%, or +21.95 points over the base model.

\paragraph{Summary of agent capabilities and limitations.}
This trace illustrates several recurring themes from the broader benchmark:

\begin{itemize}
\item \textbf{Prompt and evaluation alignment {may contribute to} early gains}: the 18$\times$ improvement from the initial SFT to the aligned version {coincided with prompt alignment and augmentation}. This echoes the broader finding that distribution mismatch between training and evaluation is often the primary bottleneck, not model capacity or training algorithm choice.
\item \textbf{Progressive strategy composition}: the agent composed SFT $\to$ GRPO $\to$ RFT, with each stage building on the previous checkpoint. This multi-stage pipeline is characteristic of successful Free-mode runs across tasks.
\item \textbf{Adaptive resource management}: the LoRA switch for GRPO demonstrates the agent's ability to diagnose and resolve engineering constraints (OOM) without human intervention.
\item \textbf{Degradation detection}: the agent detected GRPO performance regression and pivoted to RFT rather than continuing a failing strategy, showing an important metacognitive capability.
\item \textbf{Training target inconsistency from augmentation}: the mixed output format (complete function vs.\ function body) introduced conflicting supervision signals that were masked by post-processing rather than resolved at the data level.
\item \textbf{Missing ablation awareness}: the agent did not isolate the contribution of prompt alignment from data augmentation, potentially misattributing the {41.46pp} gain.
\end{itemize}

Taken together, this case shows that LLM agents can autonomously recover from severe early degradation and engineer a competitive multi-stage post-training pipeline, but their data design decisions remain brittle and their self-evaluation of \emph{why} a strategy worked is limited.

\subsection{ALFWorld on Qwen3-8B-Base}
\label{app:alfworld_trace}

The HumanEval trace above illustrates agent behavior on a static code-generation task. We now present a complementary walkthrough on ALFWorld, an interactive rollout task that is central to the benchmark's core thesis. This trace achieves {a large} absolute improvement (+88.8pp) using \emph{only} SFT, without any RL algorithm, and illustrates both the power and the limitations of agent-driven engineering on interactive tasks.

\paragraph{Setup.}
Claude Code Free mode, Qwen3-8B-Base (8.96\% baseline), 12-hour budget on 2$\times$H200. ALFWorld is a text-based household environment requiring multi-step ReAct-style interaction: the model observes the environment, reasons, executes an action, observes the result, and iterates until the task is solved or the step limit is reached. Table~\ref{tab:alfworld_trace_summary} summarizes the progression.

\begin{table}[h]
\centering
\caption{Submission progression for the ALFWorld trace walkthrough (Claude Code Free, Qwen3-8B-Base, 12h).}
\label{tab:alfworld_trace_summary}
\small
\setlength{\tabcolsep}{3pt}
\begin{tabular}{clcp{0.43\linewidth}}
\toprule
Sub \# & Script & Score & Method \\
\midrule
baseline & N/A & 8.96\% & Qwen3-8B-Base (unaligned) \\
\#1 & \texttt{train\_sft.py} & 14.93\% & SFT v1: ReAct examples \\
\#2 & \texttt{train\_sft\_v2.py} & 16.42\% & Hyperparameter tuning \\
\#4 & \texttt{train\_sft\_v4.py} & 19.40\% & Action format alignment (\texttt{put}$\to$\texttt{move}) + synthetic data \\
\#5 & \texttt{train\_sft\_v5.py} & 45.52\% & Model rollout trajectories (+26.1pp) \\
\#7 & \texttt{train\_sft\_v7.py} & \textbf{95.52\%} & Prompt format alignment (+50.0pp) \\
\#8 & \texttt{train\_sft\_v8.py} & 96.27\% & Retrain on v7 rollouts \\
\#9 & \texttt{train\_sft\_v9.py} & \textbf{97.76\%} & 4$\times$ rollout weight + all rollouts \\
\bottomrule
\end{tabular}
\end{table}

\paragraph{The hidden action-syntax trap.}
ALFWorld's original ReAct few-shot examples use the phrasing \texttt{put X in/on Y} for placement actions, but the environment's actual API requires \texttt{move X to Y}. This mismatch is not documented anywhere in the task description and must be discovered through environment interaction or code inspection. This trap creates a two-level distribution alignment problem that structures the entire trace.

\paragraph{Step 1: Baseline SFT (Sub \#1 to \#3, 14.93 to 16.42\%).}
The agent constructed training data from the 18 ReAct examples in \texttt{react\_prompts.json} (6 task types $\times$ 3 examples), using three strategies: full trajectories, permuted few-shot pairs, and step-by-step completions where the model learns to predict the next action at each step. Training used 8 epochs, lr=$2{\times}10^{-5}$, packing. The model learned to produce ReAct-format output but failed most tasks. Hyperparameter tuning (v2) and data augmentation (v3) yielded only marginal gains, indicating that the bottleneck was not in training configuration but in the data itself.

\paragraph{Step 2: Action format alignment + synthetic trajectories (Sub \#4, 19.40\%).}
The agent discovered the \texttt{put}$\to$\texttt{move} mismatch and wrote a regex-based converter applied to all training data:

\begin{promptbox}[Action Format Alignment]
\footnotesize
\begin{alltt}
def fix_put_to_move(text):
    text = re.sub(r'> put (.+?) in/on (.+)', r'> move \textbackslash{}1 to \textbackslash{}2', text)
    pattern = r'You put the (.+?) in/on the (.+?)\textbackslash{}.'
    replacement = r'You move the \textbackslash{}1 to the \textbackslash{}2.'
    text = re.sub(pattern, replacement, text)
    return text
\end{alltt}
\end{promptbox}

The agent also invested substantial effort ($\sim$200 lines) writing template-based trajectory generators for all six task types, covering diverse object and location combinations. However, this synthetic data contributed only +3pp. \textbf{Agent limitation:} the agent defaulted to ``produce more data'' before validating whether the existing data format was correct, a pattern of over-engineering before hypothesis testing.

Critically, \texttt{fix\_put\_to\_move} was applied to \emph{both} the few-shot examples and the task trajectories. This will become the root cause of the next plateau.

\paragraph{Step 3: Model rollout trajectories (Sub \#5, 45.52\%).}
The agent shifted from synthetic data to real environment interaction. Using vLLM~\citep{kwon2023efficient} to serve the v4 checkpoint, it rolled out across ALFWorld games and retained only successful trajectories:

\begin{promptbox}[Rollout Collection]
\footnotesize
\begin{alltt}
llm = LLM(model="output/v4", tensor_parallel_size=2, max_model_len=4096)

for game_no in range(num_games):
    ob, info = env.reset()
    for step in range(1, 51):
        action = llm_fn(init_prompt + history, stop=["\textbackslash{}n"])
        observation, r, done, info2 = env.step([action])
        ...
    if reward:   # keep only successful trajectories
        trajectories.append(\{"text": prompt + trajectory_text, ...\})
\end{alltt}
\end{promptbox}

This collected $\sim$264 successful trajectories, which were weighted 3$\times$ and combined with the original examples ($\sim$1,100 total samples). The result was +26pp over v4. \textbf{Real environment trajectories vastly outperformed synthetic templates}, but the score then plateaued at 45.52\% (v6 added more rollouts with no gain).

\paragraph{Step 4: Prompt format alignment, the 50-point breakthrough (Sub \#7, 95.52\%).}
The agent diagnosed the plateau by inspecting how \texttt{eval.py} constructs prompts and discovered a subtle two-level distribution mismatch:

\begin{itemize}
\item \textbf{At evaluation time}: \texttt{eval.py} uses the \emph{original} \texttt{react\_prompts.json} as few-shot examples, which contain \texttt{put X in/on Y}. The model must then output \texttt{move X to Y} as the action.
\item \textbf{At training time (v4/v5)}: \texttt{fix\_put\_to\_move()} converted \emph{everything}, including the few-shot examples, to \texttt{move}. So the model was trained on examples saying \texttt{move}, but at evaluation saw examples saying \texttt{put}.
\end{itemize}

The corrective change was to keep few-shot examples in their \emph{original} format (with \texttt{put}) while only converting the task trajectory actions to \texttt{move}, exactly matching what the model sees at evaluation time:

\begin{promptbox}[Prompt Format Alignment: v5 (Wrong) vs.\ v7 (Correct)]
\footnotesize
\begin{alltt}
# v5: examples also converted to "move" (WRONG - mismatches eval)
prompt = fix_put_to_move(react_prompts["react_put_1"])
       + fix_put_to_move(react_prompts["react_put_0"])

# v7: examples kept original (CORRECT - matches eval)
prompt = prompts["react_put_1"]     # original, with "put"
       + prompts["react_put_0"]     # original, with "put"
# Only task trajectory uses "move"
\end{alltt}
\end{promptbox}

The agent retrained from the base model (not from v5) with the corrected data (1,114 samples: 792 rebuilt rollouts + 36 original examples + 286 step-by-step completions). Training: 4 epochs, lr=$1{\times}10^{-5}$, $\sim$30 min on 2$\times$H200.

This single format change produced a +50pp jump. The alignment is counterintuitive: to produce correct behavior, the training data must preserve the ``incorrect'' phrasing in the few-shot examples, because that is what the model will see at test time.

\paragraph{Step 5: Iterative self-play (Sub \#8 to \#9, 96.27 to 97.76\%).}
The agent established a collect and retrain loop: each version's model was used to collect higher-quality rollouts, which were fed back into the next round of training.

\begin{promptbox}[Iterative Rollout Aggregation (v9)]
\footnotesize
\begin{alltt}
# Aggregate rollouts across all versions, keep shortest per game
best_by_game = \{\}
for rollout_file in ["v8.jsonl", "v7.jsonl", "v5.jsonl", "v4.jsonl"]:
    for d in load(rollout_file):
        if (
            d["game"] not in best_by_game
            or d["steps"] < best_by_game[d["game"]]["steps"]
        ):
            best_by_game[d["game"]] = d

# 271 unique games x 4 = 1084 rollouts + 36 examples
# + 286 completions = 1406 samples
\end{alltt}
\end{promptbox}

From v7 to v9, the gains were incremental (+0.75pp, +1.5pp) but consistent. The v8 collection succeeded on 271/274 games, with only 3 persistent failures on specific hard edge cases. The final 97.76\% approaches the environment's effective ceiling.

\paragraph{Comparison with other ALFWorld runs.}
This trace is one of several ALFWorld runs in the benchmark, and the comparison is instructive:

\begin{itemize}
\item \textbf{SFT-only mode} (Qwen2.5-7B-Instruct, 12h mode ablation): spent early submissions stuck at low scores before discovering the \texttt{put}$\to$\texttt{move} syntax, and \emph{never} discovered the prompt format alignment, reaching only 67.91\%. The instruction-tuned base model did not compensate for missing the deeper alignment insight.
\item \textbf{Exploratory Free mode} (Qwen2.5-7B-Instruct, 24h; not part of the main leaderboard): used environment-provided demonstrations via \texttt{extra.expert\_plan} API, collected 2,824 expert trajectories, and reached 96.27\% in 4 submissions. This is a qualitatively different strategy (demonstration collection vs.\ environment diagnosis) that nonetheless reaches a similar ceiling.
\item \textbf{Codex CLI + GPT-5.4} (Qwen3-8B-Base, 12h): reached {91.04}\% in the controlled study, showing that strong ALFWorld gains are not unique to Claude Code. The gap to the 97.76\% Claude Code result nevertheless suggests that prompt and trajectory alignment quality still separates agent stacks near the top of the task.
\end{itemize}

\paragraph{Summary of agent capabilities and limitations.}

\begin{itemize}
\item \textbf{Cross-module root-cause analysis}: the +50pp breakthrough required the agent to simultaneously understand three components, namely \texttt{eval.py}'s prompt construction, the environment's action API, and the training data pipeline, and identify the subtle inconsistency between them. This cross-module diagnostic capability is arguably the most valuable skill demonstrated across all traces.
\item \textbf{Environment interaction as data source}: the shift from synthetic templates (+3pp) to real rollouts (+26pp) demonstrates that for interactive tasks, environment interaction is not optional; it is the primary data acquisition mechanism.
\item \textbf{Iterative self-play convergence}: the collect and retrain loop (v7$\to$v8$\to$v9) produced monotonic improvements, showing that the agent can establish and execute a stable self-improvement cycle.
\item \textbf{Pure SFT sufficiency}: the entire trace uses no RL algorithm (GRPO/PPO/DPO). All 97.76\% came from SFT on increasingly better data. This supports the broader finding that agents tend to fall back to supervised pipelines, though in this case the pipeline was highly effective.
\item \textbf{Slow initial diagnosis}: the agent took 3 submissions to discover the \texttt{put}$\to$\texttt{move} mismatch, a relatively surface-level issue discoverable by running a single environment step and inspecting the returned observation. The tendency to ``train first, diagnose later'' cost exploration budget.
\item \textbf{Over-investment in low-value engineering}: the v4 synthetic trajectory generators ($\sim$200 lines covering all 6 task types) contributed only +3pp, while the real breakthrough came from a 10-line data-format alignment. The agent's default response to low scores was to produce more data rather than to question data quality.
\end{itemize}

The HumanEval and ALFWorld traces together illustrate a recurring pattern: the dominant factor in agent-driven post-training is not algorithm choice but \emph{distribution alignment}, which ensures that training data matches the evaluation protocol in format, context, and action space. Both traces' largest jumps came from aligning mismatches that were invisible in the training loss but catastrophic at evaluation time.

%% file: references.bib
@article{ouyang2022instructgpt,
  title = {Training Language Models to Follow Instructions with Human Feedback},
  author = {Ouyang, Long and Wu, Jeff and Jiang, Xu and Almeida, Diogo and Wainwright, Carroll L. and Mishkin, Pamela and Zhang, Chong and Agarwal, Sandhini and Slama, Katarina and Ray, Alex and Schulman, John and Hilton, Jacob and Kelton, Fraser and Miller, Luke and Simens, Maddie and Askell, Amanda and Welinder, Peter and Christiano, Paul and Leike, Jan and Lowe, Ryan},
  journal = {arXiv preprint arXiv:2203.02155},
  year = {2022},
  url = {https://arxiv.org/abs/2203.02155}
}

@article{deepseek2025r1,
  title = {DeepSeek-R1: Incentivizing Reasoning Capability in LLMs via Reinforcement Learning},
  author = {{DeepSeek-AI}},
  journal = {arXiv preprint arXiv:2501.12948},
  year = {2025},
  url = {https://arxiv.org/abs/2501.12948}
}

@article{cobbe2021gsm8k,
  title = {Training Verifiers to Solve Math Word Problems},
  author = {Cobbe, Karl and Kosaraju, Vineet and Bavarian, Mohammad and Chen, Mark and Jun, Heewoo and Kaiser, Lukasz and Plappert, Matthias and Tworek, Jerry and Hilton, Jacob and Nakano, Reiichiro and Hesse, Christopher and Schulman, John},
  journal = {arXiv preprint arXiv:2110.14168},
  year = {2021},
  url = {https://arxiv.org/abs/2110.14168}
}

@article{chen2021humaneval,
  title = {Evaluating Large Language Models Trained on Code},
  author = {Chen, Mark and Tworek, Jerry and Jun, Heewoo and Yuan, Qiming and Pinto, Henrique Ponde de Oliveira and Kaplan, Jared and Edwards, Harri and Burda, Yuri and Joseph, Nicholas and Brockman, Greg and Ray, Alex and Puri, Raul and Krueger, Gretchen and Petrov, Michael and Khlaaf, Heidy and Sastry, Girish and Mishkin, Pamela and Chan, Brooke and Gray, Scott and Ryder, Nick and Pavlov, Mikhail and Power, Alethea and Kaiser, Lukasz and Bavarian, Mohammad and Winter, Clemens and Tillet, Philippe and Such, Felipe Petroski and Cummings, Dave and Plappert, Matthias and Chantzis, Fotios and Barnes, Elizabeth and Herbert-Voss, Ariel and Guss, William and Nichol, Alex and Paino, Alex and Tezak, Nikolas and Tang, Jie and Babuschkin, Igor and Balaji, Suchir and Jain, Shantanu and Saunders, William and Hesse, Christopher and Carr, Andrew N. and Leike, Jan and Achiam, Josh and Misra, Vedant and Morikawa, Evan and Radford, Alec and Knight, Matthew and Brundage, Miles and Murati, Mira and Mayer, Katie and Welinder, Peter and McGrew, Bob and Amodei, Dario and McCandlish, Sam and Sutskever, Ilya and Zaremba, Wojciech},
  journal = {arXiv preprint arXiv:2107.03374},
  year = {2021},
  url = {https://arxiv.org/abs/2107.03374}
}

@article{dubois2024alpacaeval,
  title = {Length-Controlled AlpacaEval: A Simple Way to Debias Automatic Evaluators},
  author = {Dubois, Yann and Galambosi, Bal{\'a}zs and Liang, Percy and Hashimoto, Tatsunori B.},
  journal = {arXiv preprint arXiv:2404.04475},
  year = {2024},
  url = {https://arxiv.org/abs/2404.04475}
}

@article{shridhar2020alfworld,
  title = {ALFWorld: Aligning Text and Embodied Environments for Interactive Learning},
  author = {Shridhar, Mohit and Yuan, Xingdi and C{\^o}t{\'e}, Marc-Alexandre and Bisk, Yonatan and Trischler, Adam and Hausknecht, Matthew},
  journal = {arXiv preprint arXiv:2010.03768},
  year = {2020},
  url = {https://arxiv.org/abs/2010.03768}
}

@article{yao2022webshop,
  title = {WebShop: Towards Scalable Real-World Web Interaction with Grounded Language Agents},
  author = {Yao, Shunyu and Chen, Howard and Yang, John and Narasimhan, Karthik},
  journal = {arXiv preprint arXiv:2207.01206},
  year = {2022},
  url = {https://arxiv.org/abs/2207.01206}
}

@article{deepsearchqa2026,
  title = {DeepSearchQA: Bridging the Comprehensiveness Gap for Deep Research Agents},
  author = {Gupta, Nikita and Chatterjee, Riju and Haas, Lukas and Tao, Connie and Wang, Andrew and Liu, Chang and Oiwa, Hidekazu and Gribovskaya, Elena and Ackermann, Jan and Blitzer, John and Goldshtein, Sasha and Das, Dipanjan},
  journal = {arXiv preprint arXiv:2601.20975},
  year = {2026},
  url = {https://arxiv.org/abs/2601.20975}
}

@inproceedings{ross2011dagger,
  title = {A Reduction of Imitation Learning and Structured Prediction to No-Regret Online Learning},
  author = {Ross, Stephane and Gordon, Geoffrey and Bagnell, Drew},
  booktitle = {Proceedings of the Fourteenth International Conference on Artificial Intelligence and Statistics},
  pages = {627--635},
  year = {2011},
  url = {https://proceedings.mlr.press/v15/ross11a.html}
}

@article{swebench2023,
  title = {{SWE}-bench: Can Language Models Resolve Real-World {GitHub} Issues?},
  author = {Jimenez, Carlos E. and Yang, John and Wettig, Alexander and Yao, Shunyu and Pei, Kexin and Press, Ofir and Narasimhan, Karthik},
  journal = {arXiv preprint arXiv:2310.06770},
  year = {2023},
  url = {https://arxiv.org/abs/2310.06770}
}

@article{liang2022helm,
  title = {Holistic Evaluation of Language Models},
  author = {Liang, Percy and Bommasani, Rishi and Lee, Tony and Tsipras, Dimitris and Soylu, Dilara and Yasunaga, Michihiro and Zhang, Yian and Narayanan, Deepak and Wu, Yuhuai and Kumar, Ananya and Newman, Benjamin and Yuan, Binhang and Yan, Bobby and Zhang, Ce and Cosgrove, Christian and Manning, Christopher D. and R{\'e}, Christopher and others},
  journal = {arXiv preprint arXiv:2211.09110},
  year = {2022},
  url = {https://arxiv.org/abs/2211.09110}
}

@article{srivastava2022bigbench,
  title = {Beyond the Imitation Game: Quantifying and Extrapolating the Capabilities of Language Models},
  author = {Srivastava, Aarohi and Rastogi, Abhinav and Rao, Abhishek and Shoeb, Abu Awal Md and Abid, Abubakar and Fisch, Adam and Brown, Adam R. and Santoro, Adam and Gupta, Aditya and Garriga-Alonso, Adri{\`a} and others},
  journal = {arXiv preprint arXiv:2206.04615},
  year = {2022},
  url = {https://arxiv.org/abs/2206.04615}
}

@article{yang2024sweagent,
  title = {{SWE}-agent: Agent-Computer Interfaces Enable Automated Software Engineering},
  author = {Yang, John and Jimenez, Carlos E. and Wettig, Alexander and Lieret, Kilian and Yao, Shunyu and Narasimhan, Karthik and Press, Ofir},
  journal = {arXiv preprint arXiv:2405.15793},
  year = {2024},
  url = {https://arxiv.org/abs/2405.15793}
}

@article{mlebench2024,
  title = {{MLE}-bench: Evaluating Machine Learning Agents on Machine Learning Engineering},
  author = {Chan, Jun Shern and Chowdhury, Neil and Jaffe, Oliver and Aung, James and Sherburn, Dane and Mays, Evan and Starace, Giulio and Liu, Kevin and Maksin, Leon and Patwardhan, Tejal and Weng, Lilian and M{\k{a}}dry, Aleksander},
  journal = {arXiv preprint arXiv:2410.07095},
  year = {2024},
  url = {https://arxiv.org/abs/2410.07095}
}

@article{agentbench2023,
  title = {AgentBench: Evaluating {LLM}s as Agents},
  author = {Liu, Xiao and Yu, Hao and Zhang, Hanchen and Xu, Yifan and Lei, Xuanyu and Lai, Hanyu and Gu, Yu and Ding, Hangliang and Men, Kaiwen and Yang, Kejuan and Zhang, Shudan and Deng, Xiang and Zeng, Aohan and Du, Zhengxiao and Zhang, Chenhui and Shen, Sheng and Zhang, Tianjun and Su, Yu and Sun, Huan and Huang, Minlie and Dong, Yuxiao and Tang, Jie},
  journal = {arXiv preprint arXiv:2308.03688},
  year = {2023},
  url = {https://arxiv.org/abs/2308.03688}
}

@article{gaia2023,
  title = {{GAIA}: a Benchmark for General {AI} Assistants},
  author = {Mialon, Gr{\'e}goire and Fourrier, Cl{\'e}mentine and Swift, Craig and Wolf, Thomas and LeCun, Yann and Scialom, Thomas},
  journal = {arXiv preprint arXiv:2311.12983},
  year = {2023},
  url = {https://arxiv.org/abs/2311.12983}
}

@inproceedings{huang2024mlagentbench,
  title = {{MLAgentBench}: Evaluating Language Agents on Machine Learning Experimentation},
  author = {Huang, Qian and Vora, Jian and Liang, Percy and Leskovec, Jure},
  booktitle = {International Conference on Machine Learning (ICML)},
  year = {2024},
  url = {https://openreview.net/forum?id=1Fs1LvjYQW}
}

@inproceedings{wang2024openhands,
  title = {{OpenHands}: An Open Platform for {AI} Software Developers as Generalist Agents},
  author = {Wang, Xingyao and Li, Boxuan and Song, Yufan and Xu, Frank F. and Tang, Xiangru and Zhuge, Mingchen and Pan, Jiayi and Song, Yueqi and Li, Bowen and Singh, Jaskirat and Tran, Hoang H. and Li, Fuqiang and Ma, Ren and Zheng, Mingzhang and Qian, Bill and Shao, Yanjun and Muennighoff, Niklas and Zhang, Yizhe and Hui, Binyuan and Lin, Junyang and Brennan, Robert and Peng, Hao and Ji, Heng and Neubig, Graham},
  booktitle = {The Thirteenth International Conference on Learning Representations (ICLR)},
  publisher = {OpenReview.net},
  year = {2025},
  url = {https://openreview.net/forum?id=OJd3ayDDoF}
}

@inproceedings{yao2023react,
  title = {{ReAct}: Synergizing Reasoning and Acting in Language Models},
  author = {Yao, Shunyu and Zhao, Jeffrey and Yu, Dian and Du, Nan and Shafran, Izhak and Narasimhan, Karthik and Cao, Yuan},
  booktitle = {International Conference on Learning Representations (ICLR)},
  year = {2023},
  url = {https://openreview.net/forum?id=WE_vluYUL-X}
}

@article{wijk2024rebench,
  title = {{RE-Bench}: Evaluating Frontier {AI} {R\&D} Capabilities of Language Model Agents against Human Experts},
  author = {Wijk, Hjalmar and Lin, Tao and Becker, Joel and Jawhar, Sami and Parikh, Neev and Broadley, Thomas and Chan, Lawrence and Chen, Michael and Clymer, Josh and Dhyani, Jai and Ericheva, Elena and Garcia, Katharyn and Goodrich, Brian and Jurkovic, Nikola and Karnofsky, Holden and Kinniment, Megan and Lajko, Aron and Nix, Seraphina and Sato, Lucas and Saunders, William and Taran, Maksym and West, Ben and Barnes, Elizabeth},
  journal = {arXiv preprint arXiv:2411.15114},
  year = {2024},
  url = {https://arxiv.org/abs/2411.15114}
}

@article{nathani2025mlgym,
  title = {{MLGym}: A New Framework and Benchmark for Advancing {AI} Research Agents},
  author = {Nathani, Deepak and Madaan, Lovish and Roberts, Nicholas and Bashlykov, Nikolay and Menon, Ajay and Moens, Vincent and Budhiraja, Amar and Magka, Despoina and Vorotilov, Vladislav and Chaurasia, Gaurav and Hupkes, Dieuwke and Cabral, Ricardo Silveira and Shavrina, Tatiana and Foerster, Jakob and Bachrach, Yoram and Wang, William Yang and Raileanu, Roberta},
  journal = {arXiv preprint arXiv:2502.14499},
  year = {2025},
  url = {https://arxiv.org/abs/2502.14499}
}

@inproceedings{posttrainbench2026,
  title = {PostTrainBench: Can {LLM} Agents Automate {LLM} Post-Training?},
  author = {Rank, Ben and Bhatnagar, Hardik and Prabhu, Ameya and Eisenberg, Shira and Nguyen, Karina and Bethge, Matthias and Andriushchenko, Maksym},
  booktitle = {International Conference on Machine Learning (ICML)},
  year = {2026},
  url = {https://arxiv.org/abs/2603.08640}
}

@article{webarena2023,
  title = {{WebArena}: A Realistic Web Environment for Building Autonomous Agents},
  author = {Zhou, Shuyan and Xu, Frank F. and Zhu, Hao and Zhou, Xuhui and Lo, Robert and Sridhar, Abishek and Cheng, Xianyi and Ou, Tianyue and Bisk, Yonatan and Fried, Daniel and Alon, Uri and Neubig, Graham},
  journal = {arXiv preprint arXiv:2307.13854},
  year = {2023},
  url = {https://arxiv.org/abs/2307.13854}
}

@article{rafailov2023dpo,
  title = {Direct Preference Optimization: Your Language Model is Secretly a Reward Model},
  author = {Rafailov, Rafael and Sharma, Archit and Mitchell, Eric and Ermon, Stefano and Manning, Christopher D. and Finn, Chelsea},
  journal = {arXiv preprint arXiv:2305.18290},
  year = {2023},
  url = {https://arxiv.org/abs/2305.18290}
}

@article{shao2024deepseekmath,
  title = {DeepSeekMath: Pushing the Limits of Mathematical Reasoning in Open Language Models},
  author = {Shao, Zhihong and Wang, Peiyi and Zhu, Qihao and Xu, Runxin and Song, Junxiao and Bi, Xiao and Zhang, Haowei and Zhang, Mingchuan and Li, Y.K. and Wu, Y. and Guo, Daya},
  journal = {arXiv preprint arXiv:2402.03300},
  year = {2024},
  url = {https://arxiv.org/abs/2402.03300}
}

@article{li2026ftdojo,
  title = {{FT-Dojo}: Towards Autonomous {LLM} Fine-Tuning with Language Agents},
  author = {Li, Qizheng and Zhang, Yifei and Yang, Xiao and Yang, Xu and Wang, Zhuo and Liu, Weiqing and Bian, Jiang},
  journal = {arXiv preprint arXiv:2603.01712},
  year = {2026},
  url = {https://arxiv.org/abs/2603.01712}
}

@article{qwen2024qwen25,
  title = {{Qwen2.5} Technical Report},
  author = {{Qwen Team}},
  journal = {arXiv preprint arXiv:2412.15115},
  year = {2024},
  url = {https://arxiv.org/abs/2412.15115}
}

@article{qwen2025qwen3,
  title = {{Qwen3} Technical Report},
  author = {{Qwen Team}},
  journal = {arXiv preprint arXiv:2505.09388},
  year = {2025},
  url = {https://arxiv.org/abs/2505.09388}
}

@inproceedings{kwon2023efficient,
  title = {Efficient Memory Management for Large Language Model Serving with {PagedAttention}},
  author = {Kwon, Woosuk and Li, Zhuohan and Zhuang, Siyuan and Sheng, Ying and Zheng, Lianmin and Yu, Cody Hao and Gonzalez, Joseph E. and Zhang, Hao and Stoica, Ion},
  booktitle = {Proceedings of the ACM SIGOPS 29th Symposium on Operating Systems Principles},
  year = {2023},
  url = {https://arxiv.org/abs/2309.06180}
}
